\documentclass[10pt,twocolumn,letterpaper]{article}

\usepackage[pagenumbers]{cvpr} 

\usepackage[dvipsnames]{xcolor}

\usepackage{times}
\usepackage{epsfig}
\usepackage{graphicx}
\usepackage{amsmath}
\usepackage{amssymb}
\usepackage{color,xcolor}

\usepackage{comment}

\usepackage{pifont}

\usepackage{microtype}
\usepackage[font=small]{caption}
\usepackage{arydshln}
\usepackage{tabularx}
\usepackage{adjustbox}
\usepackage{array}
\usepackage{booktabs}
\usepackage{colortbl}
\usepackage{float,wrapfig}
\usepackage{hhline}
\usepackage{multirow}
\usepackage{subcaption} %
\usepackage[percent]{overpic}

\usepackage{stmaryrd}

\usepackage{amsmath,amsfonts,amssymb}
\usepackage{bm}
\usepackage{nicefrac}
\usepackage{microtype}
\usepackage{dsfont}
\usepackage{changepage}
\usepackage{extramarks}
\usepackage{fancyhdr}
\usepackage{setspace}
\usepackage{soul}
\usepackage{xspace}

\usepackage{algorithm, algorithmic}
\usepackage{enumitem}
\usepackage[title]{appendix}

\usepackage{duckuments}
\usepackage{subcaption}

\usepackage{fp}
\usepackage{expl3}[2012-07-08]
\ExplSyntaxOn

\ExplSyntaxOff

\usepackage{amsmath,amsfonts,bm}

\def\x{{x}}
\def\z{{z}}

\def\xi{{\x_i}}

\newcommand{\lblfig}[1]{\label{fig:#1}}
\newcommand{\lblsec}[1]{\label{sec:#1}}
\newcommand{\lbleq}[1]{\label{eq:#1}}
\newcommand{\lbltbl}[1]{\label{tbl:#1}}

\newcommand{\ignorethis}[1]{}

\def\eqref#1{equation~\ref{#1}}

\def\1{\bm{1}}

\def\eps{{\epsilon}}

\def\vc{{\bm{c}}}

\def\vu{{\bm{u}}}
\def\vv{{\bm{v}}}
\def\vw{{\bm{w}}}
\def\vx{{\bm{x}}}

\def\vz{{\bm{z}}}
\def\veps{{\bm{\eps}}}

\DeclareMathAlphabet{\mathsfit}{\encodingdefault}{\sfdefault}{m}{sl}
\SetMathAlphabet{\mathsfit}{bold}{\encodingdefault}{\sfdefault}{bx}{n}

\newcommand{\ignore}[1]{}

\makeatletter
\DeclareRobustCommand\onedot{\futurelet\@let@token\@onedot}
\def\@onedot{\ifx\@let@token.\else.\null\fi\xspace}

\makeatother

\definecolor{MyDarkBlue}{rgb}{0,0.08,1}
\definecolor{MyDarkGreen}{rgb}{0.02,0.6,0.02}
\definecolor{MyDarkRed}{rgb}{0.8,0.02,0.02}
\definecolor{MyDarkOrange}{rgb}{0.40,0.2,0.02}
\definecolor{MyPurple}{RGB}{111,0,255}
\definecolor{MyRed}{rgb}{1.0,0.0,0.0}
\definecolor{MyGold}{rgb}{0.75,0.6,0.12}
\definecolor{MyDarkgray}{rgb}{0.66, 0.66, 0.66}
\definecolor{myorange}{RGB}{255,69,0}

\definecolor{revision}{RGB}{255,69,0}

\newcommand{\model}{Control3Diff}

\definecolor{cvprblue}{rgb}{0.21,0.49,0.74}
\usepackage[pagebackref,breaklinks,colorlinks,citecolor=cvprblue]{hyperref}

\def\paperID{299} 
\def\confName{3DV\xspace}
\def\confYear{2024\xspace}

\title{Control3Diff: Learning Controllable 3D Diffusion Models\\ from Single-view Images}
\author{
Jiatao Gu\textsuperscript{$\dag$},
Qingzhe Gao\textsuperscript{$\S$},
Shuangfei Zhai\textsuperscript{$\dag$},
Baoquan Chen\textsuperscript{$\P$},
Lingjie Liu\textsuperscript{$\ddag$},
Josh Susskind\textsuperscript{$\dag$}
\\
\textsuperscript{$\dag$}Apple \quad 
\textsuperscript{$\ddag$}University of Pennsylvania \quad 
\textsuperscript{$\S$}Shandong University \quad 
\textsuperscript{$\P$}Peking University \\
{\tt\small 
\textsuperscript{$\dag$}\{jgu32,szhai,jsusskind\}@apple.com \quad 
\textsuperscript{$\ddag$} lingjie.liu@seas.upenn.edu} \\ 
{\tt \small \textsuperscript{$\S$} gaoqingzhe97@gmail.com 
\quad \textsuperscript{$\P$} baoquan@pku.edu.cn 
}
}

\begin{document}
\twocolumn[{%
\maketitle
\centering
\vspace{-17pt}
\includegraphics[width=\linewidth]{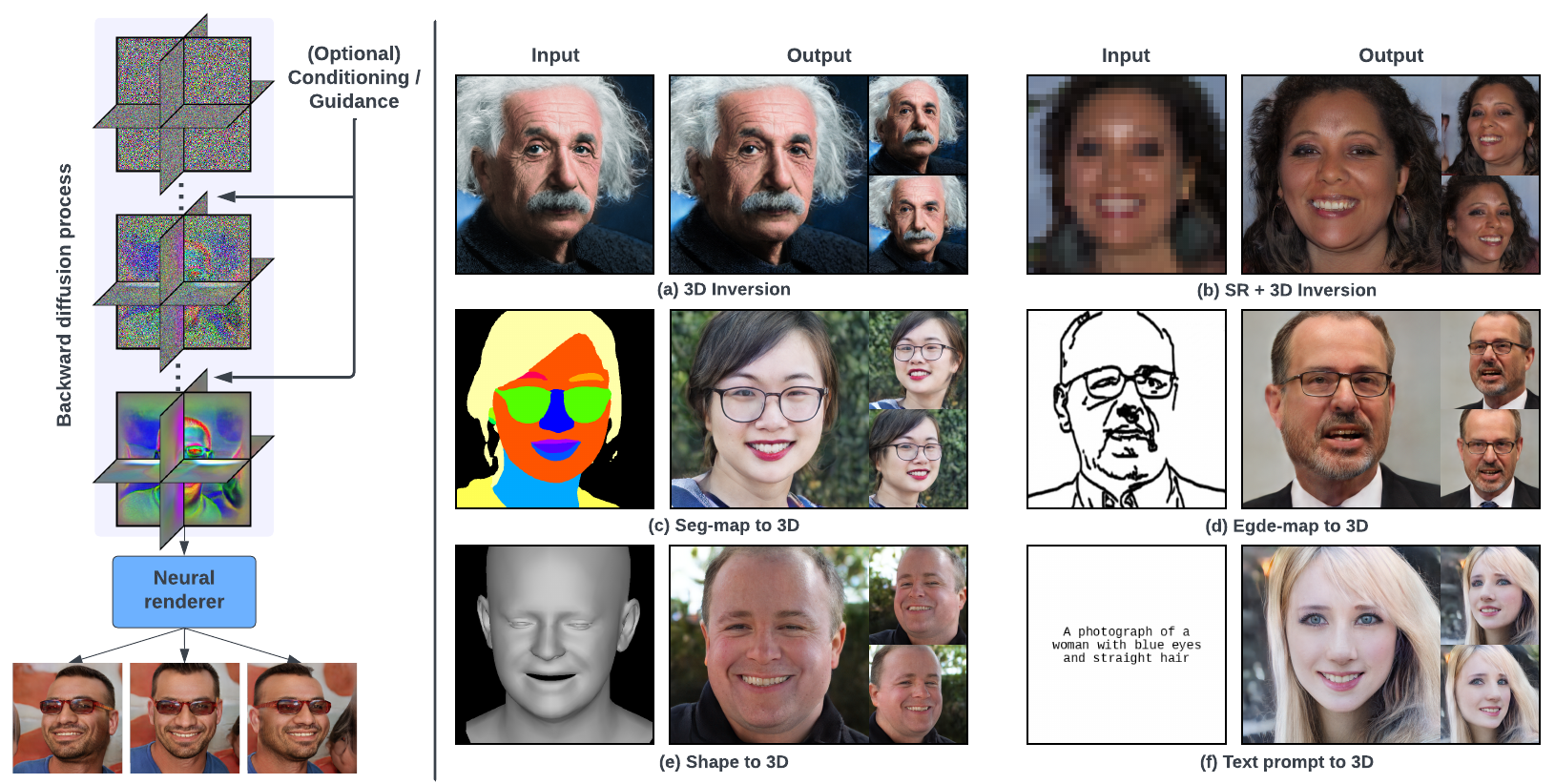}
 \vspace{-1.5em}
\captionof{figure}{Left is the generation process, where a diffusion model samples a triplane which can be used for image rendering. Right are the examples of controllable generation given various conditioning inputs, showing generated frontal and side views from {\model}. The faces shown are all generated by models without real identities \textbf{due to concerns about individual consent} except for the input in (a).
}
 \vspace{2em}
\lblfig{teaser}
}]
\begin{abstract}\vspace{-6pt}
Diffusion models have recently become the de-facto approach for generative modeling in the 2D domain. However, extending diffusion models to 3D is challenging, due to the difficulties in acquiring 3D ground truth data for training.  
On the other hand, 3D GANs that integrate implicit 3D representations into GANs have shown remarkable 3D-aware generation when  trained only on single-view image datasets. 
However, 3D GANs do not provide straightforward ways to precisely control image synthesis. 
To address these challenges, 
We present {\model}, a 3D diffusion model that combines the strengths of diffusion models and 3D GANs for versatile controllable 3D-aware image synthesis for single-view datasets. 
{\model} explicitly models the underlying latent distribution (optionally conditioned on external inputs), thus enabling direct control during the diffusion process. Moreover, our approach is general and applicable to any types of controlling inputs, allowing us to train it with the same diffusion objective without any auxiliary supervision. We validate the efficacy of {\model} on standard image generation benchmarks including FFHQ, AFHQ, and ShapeNet, using various conditioning inputs such as images, sketches, and text prompts. 
\end{abstract}

\section{Introduction}
The synthesis of photo-realistic 3D-aware images of real-world scenes from sparse controlling inputs is a long-standing problem in both computer vision and computer graphics, with various applications including robotics simulation, gaming, and virtual reality. Depending on the task, sparse inputs can be single-view images, guiding poses, or text instructions, and the objective is to recover 3D representations and synthesize consistent images from novel viewpoints. This is a challenging problem, as the sparse inputs typically contain insufficient information to predict complete 3D details. Consequently, the selection of an appropriate \textit{prior} during controllable generation is crucial for resolving uncertainties.
Recently, significant progress has been made in the field of 2D image generation through the use of diffusion-based generative models~\cite{sohl2015deep,ho2020denoising,song2019generative,dhariwal2021diffusion}, which learn the \textit{prior} and have achieved remarkable success in various conditional applications such as super-resolution~\cite{saharia2021image,li2022srdiff,gu2022f}, in-painting~\cite{lugmayr2022repaint}, image translation~\cite{saharia2022palette,zhang2023adding}  and text-guided synthesis~\cite{ramesh2022hierarchical, rombach2021highresolution,saharia2022photorealistic,ho2022imagen}.  It is natural to consider applying similar approaches in 3D generation. However, learning diffusion models typically relies heavily on the availability of ground-truth data, which is not commonly available for 3D content, especially for single-view images.

To address this limitation, we propose a framework called {\model}, which links diffusion models to generative adversarial networks (GANs)~\cite{Goodfellow2014} and takes advantage of the success of GANs in 3D-aware image synthesis~\cite{schwarz2020graf,chanmonteiro2020pi-GAN,giraffe,gu2021stylenerf,eg3d,or2022stylesdf,epigraf}. 
The core idea behind 3D GANs is to learn a generator based on neural fields that fuse 3D inductive bias in modeling with volume rendering. By training 3D GANs on single-view data with random noises and viewpoints as inputs, we can avoid the need for 3D ground truth. Our proposed framework {\model} predicts the internal states of 3D GANs given any conditioning inputs by modeling the prior distribution of the underlying manifolds of real data using diffusion models. Furthermore, the proposed framework can be trained on synthetic generation from a 3D GAN, allowing for infinite examples to be used for training without worrying about over-fitting. Finally, by applying the guidance techniques~\cite{dhariwal2021diffusion,ho2022classifier} in 2D diffusion models, we are able to learn controllable 3D generation with a single loss function for all conditional tasks. This eliminates the use of ad-hoc supervisions which were commonly needed in existing conditional 3D generation~\cite{pix2nerf,deng20233d}. 

To validate the proposed framework, we use a variant of the recently proposed EG3D~\cite{eg3d} that learns an efficient tri-plane representation as the basis for {\model}. We extensively conduct experiments on six types of inputs and demonstrate the effectiveness of {\model} on standard benchmarks including FFHQ, AFHQ-cat, and ShapeNet.

\section{Preliminaries: Controllable Image Synthesis}
In this section, we first define the problem of \textit{controllable image synthesis in 2D and 3D-aware manners} and review the 2D solutions with diffusion models. Then, we pose the difficulties of applying similar methods to 3D scenario.
\subsection{Definition}
\lblsec{definition}
\paragraph{2D.} The goal of controllable synthesis is to learn a generative model that synthesizes diverse 2D images $\vx$ conditioned on an input control signal $\vc$.  This can be mainly done by sampling images in the following two ways:  
\begin{equation}
   \vx \sim \exp\left[-\ell(\vc,\vx)\right]\cdot p_\theta(\vx) \quad \text{Or}  \quad  p_\theta(\vx | \vc),
   \lbleq{control_2d}
\end{equation}
where $\theta$ is the parameters of the generative model.
The former one is called \emph{guidance}. At test-time, an energy function $\ell(\vc,\vx)$ is to measure the alignment between the synthesized image $\vx$ and the input $\vc$ to guide the prior generation $p_\theta(\vx)$. Note that, only for the controllable tasks where the energy function $\ell(\vc,\vx)$ can be defined, the \emph{guidance} techniques can be applied. 
The latter one directly formulates it as a \emph{conditional} generation problem $p_\theta(\vx|\vc)$ if the paired data $(\vx, \vc)$  is available. As $\vc$ typically contains less information than $\vx$, it is crucial to handle uncertainties with  generative models.
\vspace{-10pt}
\paragraph{3D.}The above formulation can be simply extended to 3D. In this work, we assume a 3D scene represented by latent representation $\vz$, and we synthesize 3D-consistent images by rendering $\vx=\mathcal{R}(\vz,\pi)$ given different camera poses $\pi$. Here, we do not restrict the space of $\vz$, meaning that it can be any high-dimensional structure that describes the 3D scene.
Similarly, we can define 3D-aware controllable image synthesis by replacing $\vx$ with $\vz$ in \cref{eq:control_2d}.


\subsection{Diffusion Models}
\lblsec{diffusion}
Standard diffusion models~\cite{sohl2015deep,song2019generative,ho2020denoising} are explicit generative models defined by a Markovian process. 
Given an image $\vx$, a diffusion model defines continuous time latent variables $\{\vz_t | t\in [0, 1], \vz_0=\vx\}$ based on a fixed schedule $\{\alpha_t, \sigma_t\}$:
$    q(\vz_t | \vz_s) = \mathcal{N}(\vz_t; \alpha_{t|s}\vz_s, \sigma^2_{t|s}I),  \ \ 0 \leq s < t \leq 1,$
where $\alpha_{t|s} = \alpha_t/\alpha_s, \sigma^2_{t|s}=\sigma_t^2-\alpha_{t|s}^2\sigma_s^2$.
Following this definition, we can easily derive the latent $\vz_t$ at any time by 
$q(\vz_t | \vz_0) = \mathcal{N}(\vz_t;\alpha_t\vz_0, \sigma_t^2I).$ 
The model $\theta$ then learns the reverse process by denoising $\vz_t$ to the clean target $\vx$ with a weighted reconstruction loss $\mathcal{L}_\theta$: 
\begin{equation}
    \mathcal{L}_{\text{Diff}} = \mathbb{E}_{\vz_t\sim q(\vz_t | \vz_0), t \sim [0, 1]}\left[\omega_t\cdot\|\vz_\theta(\vz_t) - \vz_0\|_2^2\right].
    \lbleq{diff_loss}
\end{equation}
Typically, $\theta$ is parameterized as a U-Net~\cite{ronneberger2015,ho2020denoising} or ViT~\cite{peebles2022scalable}. 
Sampling from a learned model $p_\theta$ can be performed using ancestral sampling rules~\cite{ho2020denoising} 
-- staring with pure Gaussian noise $\vz_1\sim\mathcal{N}(0,I)$, we sample $s, t$ following a uniformly spaced sequence from 1 to 0: 
\begin{equation}
    \vz_s=\alpha_s\vz_\theta(\vz_t) + \sqrt{\sigma^2_s-\bar{\sigma}^2}\veps_\theta(\vz_t) + \bar{\sigma}\veps,  \veps\sim\mathcal{N}(0,I),
    \lbleq{diff_sample}
\end{equation} 
where $\bar{\sigma}=\sigma_s\sigma_{t|s}/\sigma_t$ and $\veps_\theta(\vz_t) = (\vz_t - \alpha_t \vz_\theta(\vz_t)) / \sigma_t$.
By decomposing the sophisticated generative process into hundreds of denoising steps, diffusion models effectively expand the modeling capacity, and have been shown superior performance than other types of generative models~\cite{dhariwal2021diffusion}.
For better efficiency, Latent Diffusion Models (LDM~\cite{rombach2021highresolution}) have extended the process in latent space by learning an additional encoder $\vz_0=\mathcal{E}(\vx)$ to map input images to the latent space. 

Due to the autoregressive nature, diffusion models are suitable for controllable generation (\cref{sec:definition}). Prior research studied \textit{guidance} with a variety of classifiers, constraints or auxiliary loss functions~\cite{dhariwal2021diffusion,kawar2022denoising,chung2022diffusion,graikos2022diffusion,ho2022video,gu2023nerfdiff,bansal2023universal}. Other works explored learning \textit{conditional diffusion} models with parallel data (e.g., class labels~\cite{ho2022classifier}, text prompts~\cite{rombach2021highresolution}, aligned image maps~\cite{zhang2023adding}). 
Importantly, \textit{classifier-free guidance}~\cite{ho2022classifier}, which enables generation with a balance between sampling controlled quality and diversity, has become a basic building block for large-scale diffusion models~\cite{saharia2022photorealistic,ramesh2022hierarchical}.

\begin{figure*}[thb]
\begin{center}
    \includegraphics[width=\linewidth]{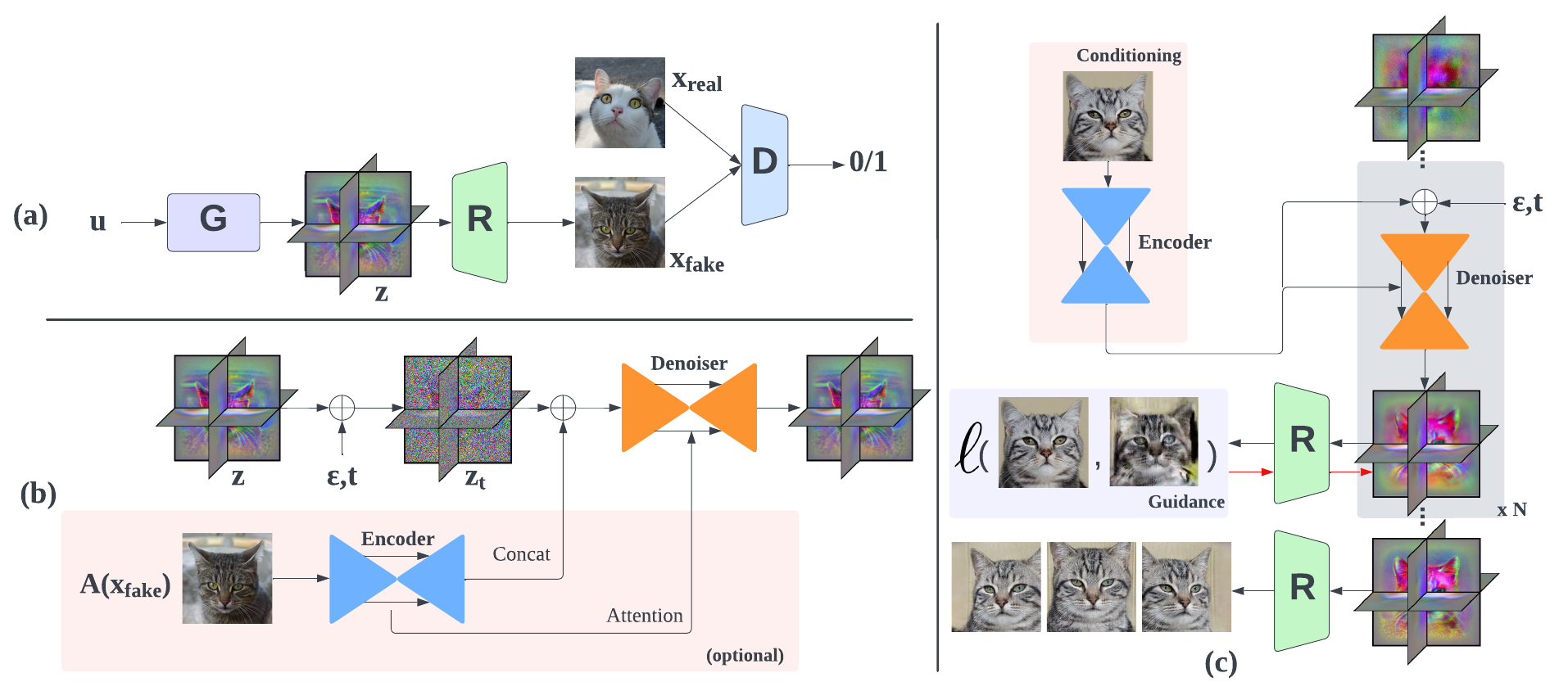}
\end{center}
    \vspace{-10pt}
   \caption{Pipeline of {\model}. (a) 3D GAN training; (b) Diffusion model trained on the extracted tri-planes can be trained with or without the input conditioning; (c) controllable 3D generation with the learned diffusion model, optionally with guidance. The tri-plane features are presented in three color planes, and the camera poses are omitted for better visual convenience.}
   \vspace{-10pt}
\lblfig{pipeline}
\end{figure*}

\subsection{3D-aware Image Synthesis}
When extending image synthesis to 3D, one can model each 3D scene, which corresponds to a latent representation $\vz$ (\cref{sec:definition}), as a neural radiance field (NeRF~\cite{mildenhall2020nerf}) $f_\vz: \mathbb{R}^5\rightarrow \mathbb{R}_+^4$ which maps every spatial point and the viewing direction to its radiance and density. $f_\vz$ is parameterized as MLPs~\cite{mildenhall2020nerf} or tri-planes~\cite{eg3d,tensorf} with upsamplers~\cite{giraffe,gu2021stylenerf}. 
Next, we can synthesize 3D-consistent images 
via volume rendering~\cite{max}.
 
Despite the success in the 2D scenario, diffusion models have rarely been applied directly in controllable 3D-aware image synthesis with NeRF. There are three \textbf{key challenges}: 
 \begin{enumerate}[leftmargin=*]
     \item Learning diffusion models requires the 3D ground-truth $\vz$ (shown in \cref{eq:diff_loss}) that is often unavailable.
    \item While there exist approaches to acquire high-quality 3D labels from dense multi-view image collections, for most of the cases, only single-view images are available. 
     \item As discussed in~\cref{sec:definition}, we need either an energy function $\ell(.,.)$ for \emph{guidance} or paired data for \emph{conditional generation} in controllable synthesis. However, both of them are not straightforward to define in the latent space of implicit 3D representations (e.g., NeRF).
 \end{enumerate}
More precisely, targeting on Challenge \# 1, prior arts~\cite{bautista2022gaudi,shue20223d,muller2022diffrf,wang2022rodin} first reconstruct the latent $\vz$ from dense multi-view images of each scene. In the rest of paper, we refer to them as \textit{reconstruction-based} methods. That is, given a set of posed images $\{\vx_i\}_{i=1}^N$, one can minimize:
\begin{equation}
    \mathcal{L}_{\text{RC}} = \mathbb{E}_{\{\vx_i\}\sim\text{data}}\left[\sum_{i}\|\mathcal{R}(\vz, \pi_i) - \vx_i\|^2_2 + \mathcal{H}(\vz)\right],
    \vspace{-5pt}
    \lbleq{recon}
\end{equation}
where $\pi_i$ is the camera of $\vx_i$, $\mathcal{R}$ is the differentiable volume renderer of $f_\vz$, and $\mathcal{H}$ is the prior regularization over $\vz$. Here, $\z=\mathcal{E}(\{\vx_i\}_{i=1}^N)$ represents either the backward process of $\nabla_\vz\mathcal{L}_{\text{recon}}$ (also known as ``auto-decoder''~\cite{Sitzmann2019}) that updates $\vz$ via SGD, or an amortized multi-view encoder~\cite{nerfvae,srt22}. 

In spite of the good results with dense multi-view data for training, 
these methods perform poorly when only one view is available for each scene (Challenge \#2). Single-view auto-decoders usually fail to learn geometry, even with strong regularization~\cite{rebain2022lolnerf},  the reconstructed quality is still limited. On the other side, using an encoder $\mathcal{E}(\vx)$ may ease the aforementioned issues after adopting various auxiliary losses with novel-view rendering~\cite{pix2nerf}. Yet, due to limited view coverage and object occlusion, an image encoder is unable to predict fully determined 3D details, resulting in additional uncertainties. 
Besides \#1 and \#2, \#3 has rarely been studied in prior research. In the next section, we will elaborate on how we address these challenges to achieve controllable 3D-aware image synthesis with only single-view images for training. 

\section{Method: {\model}}
We present {\model}, a controllable 3D-aware generation framework based on a 3D GAN (\cref{sec:model}). 
We study two ways of controlling image synthesis with {\model} (\cref{sec:guidance,sec:conditional}).
The pipeline is illustrated in \cref{fig:pipeline}.
\subsection{Latent Diffusion with 3D GANs}
\lblsec{model}
Instead of acquiring $\vz$ from dense multi-view images $\{\vx_i\}$ as done in reconstruction-based methods, we directly sample from the learned distribution of $\vz$ of a 3D GAN model, which is trained on single-view images. In this paper, considering its state-of-the-art performance, we build {\model} based on EG3D~\cite{eg3d}.
EG3D first learns a tri-plane generator $\mathcal{G}: \vu\in \mathbb{R}^{512} \rightarrow \vz \in \mathbb{R}^{3\times 256\times 256\times 32}$, mapping low-dimensional noises to an expressive  tri-plane. The feature of a 3D point is obtained by projecting the point to three orthogonal planes and gathering local features from the three planes, which is the input to the radiance function $f_\vz$ for radiance and density prediction.

Training an EG3D model requires a joint optimization of a camera-conditioned discriminator $\mathcal{D}$, and we adopt the non-saturating logistic objective with R1 regularization:
\begin{equation}
    \begin{split}
    \mathcal{L}_{\text{GAN}}&=\mathbb{E}_{\vu\sim \mathcal{N}(0,I),\pi\sim \Pi}\left[h\left(\mathcal{D}(
    \mathcal{R}\left(
        \mathcal{G}(\vu), \pi\right), \pi\right)\right] \\+ 
        &\mathbb{E}_{\vx,\pi\sim\text{data}}\left[h\left(-\mathcal{D}(\vx, \pi)\right) + \gamma\|\nabla_\vx\mathcal{D}(\vx,\pi)\|^2_2\right],
    \end{split}
\lbleq{gan_loss}
\end{equation}
where $h=-\log(1+\exp(-u))$ and $\Pi$ is the prior camera distribution. 
The adversarial learning enables the training on single-view images, as it only forces the model output to match the training data distribution rather than learns a one-to-one mapping as an auto-encoder. 
Note that, in order to train diffusion models more stable, besides \cref{eq:gan_loss}, we also bound $\mathcal{G}(\vu)$ with $\tanh(.)$ and apply an additional L2 loss similar to~\cite{shue20223d} when training EG3D. However, we observed in our experiments that these additional constraints would not affect the performance of EG3D.

After EG3D is trained, as the second stage, we apply the denoising on the tri-plane to train a diffusion model 
with the  renderer $\mathcal{R}$ frozen.  
Training follows the same denoising objective~\cref{eq:diff_loss} and $\vz_0=\mathcal{G}(\vu)$.
As $\vu$ is randomly sampled, we can essentially learn from unlimited data. 
Different from~\cite{muller2022diffrf,wang2022rodin}, we do not need any auxiliary loss or  architectural change. Optionally, we can add the control signal as the conditioning to the diffusion network to formulate a \textit{conditional} generation framework (\cref{sec:conditional}).    

We note that, training a diffusion model over $\mathcal{G}(\vu)$ samples differs from distilling a pre-trained GAN into another GAN generator, which is unsuitable for the controlling tasks. 
Although it is efficient to sample high-quality tri-planes $\vz$, GANs are implicit generative models~\cite{mohamed2016learning} and do not model the likelihood in the learned latent space. 
That is to say, we do not have a proper prior $p(\vz)$ given the latent representations of a 3D GAN. Especially in the high-dimensional space like tri-planes, any control without knowing the underlying density will easily fall off the learned manifold and output degenerated results. 
As a result, almost all existing works~\cite{pix2nerf,deng20233d} utilize 3D GANs for controlling the focus on low-dimensional spaces, which can be approximately assumed Gaussian. However, this has to scarify the controllability.
In contrast, diffusion models explicitly learn the score functions of the latent distribution even with high-dimensionality~\cite{song2019generative}, which fills in the missing pieces for 3D GANs. 
Also see experimental comparison in~\cref{table:inversion}.

\subsection{Conditional 3D Diffusion}
\lblsec{conditional}  
We can synthesize controllable images by extending latent diffusion into a conditional generation framework.
Conventionally, learning such conditional models requires labeling parallel corpus, e.g., large-scale text-image pairs~\cite{schuhmann2022laion} for the Text-to-Image task. 
Compared to acquiring 2D paired data, creating the paired data of the control signal and 3D representation is much more difficult. 
In our method, however, we can easily synthesize an infinite number of pairs of the control signal and triplanes by using the rendered images of the triplane from 3D GAN to predict the control signal with an off-the-shelf method. Now, the learning objective can be written as follows:
\begin{equation}    \mathcal{L}_{\text{Cond}}=\mathbb{E}_{\vz_0,\vz_t, t,\pi}\left[\omega_t\cdot\|\vz_\theta(\vz_t, \mathcal{A}\left(\mathcal{R}(\vz_0, \pi)\right)) - \vz_0\|_2^2\right].
    \lbleq{cond_diff}
\end{equation}
where $\vz_0=\mathcal{G}(\vu)$ is the sampled tri-plane, $\mathcal{A}$ is the off-the-shelf prediction module that converts rendered images into $\vc$ (e.g., ``edge-detector'' for edge-map to 3D generation), and
$\pi\sim\Pi$ is a pre-defined camera distribution based on the testing preference. Here $\vz_\theta$ represents a conditional denoiser that learns to predict denoised tri-planes given the condition. 
In early exploration, we noticed that the prior camera distribution $\Pi$ significantly impacts the generalizability of the learned model, where for some datasets (e.g., FFHQ, AFHQ), the biased camera distribution in training set would cause degenerated results for rare camera views. Therefore, we specifically re-sample the cameras for these datasets. 

\begin{figure*}[t]
    \centering
    \includegraphics[width=\linewidth]{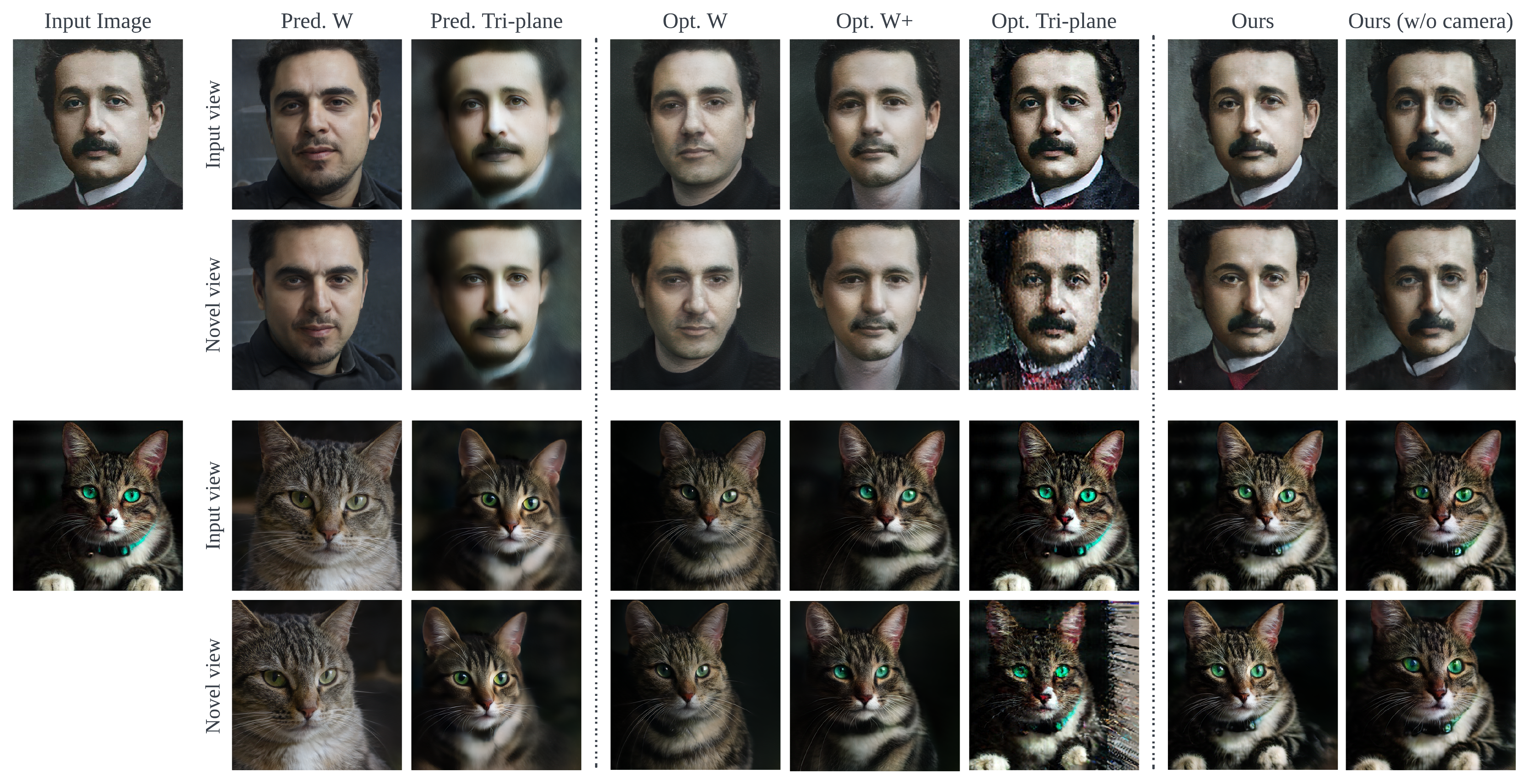}
    \caption{Comparison for 3D-inversion of \textit{in-the-wild} images. 
    We compare the proposed approach to direct prediction of the GAN's latent $\mathcal{W}$ and Tri-plane with a learned encoder, as well as an optimization based approach to infer the latent and expanded latent $\mathcal{W}$, $\mathcal{W+}$, as well as the Tri-plane, following \cite{abdal2020image2stylegan++}. Our method achieves better view consistency with higher output image quality compared to baselines.}
    
    \vspace{-5pt}
    \lblfig{compare_invert}
\end{figure*}
\paragraph{Joint Diffusion with Camera Pose $\pi_c$}
Conditional models can be learned without camera input, which implicitly maps the input view to the global triplane space. 
It implies that conditional models can predict camera information through diffusion.
In light of this observation, we propose to jointly predict the input camera pose $\pi_c$ with $\vz$ in one diffusion framework. Similar to 3D-aware generation, predicting cameras from a single view is also a challenging problem, requiring resolving ambiguities in natural images. At the same time, previous works either rely on external deterministic camera predictors~\cite{lin20223d} or optimize the cameras at inference time~\cite{ko20233d}. In this work, for simplicity, we flatten $\pi_c$ into a vector, broadcast it, and concatenate it to the channels of $\vz$ as the new diffusion target.

\subsection{Guided 3D Diffusion}
\lblsec{guidance}
An alternative way to control image synthesis is to follow a similar recipe in 2D (as defined in \cref{sec:definition}) to perform test-time guidance based on a task-specific energy function $\ell(\vc, \vz)$. Nevertheless, directly defining such an energy function between $\vc$ and 3D representation (i.e., a tri-plane NeRF) is challenging. 
We circumvent this by defining $\ell(.,.)$ to measure the closeness between $\vc$ and the differentiablly rendered image $\mathcal{R}(\vz, \pi_c)$. 
In this way, we can learn the 3D representation using 2D rendering guidance (e.g., CLIP score~\cite{radford2021learning} for text-to-3D, and MSE or perceptual loss~\cite{Zhang2018f} for image inversion). 
Using 2D guidance for learning 3D representation is reasonable since the final targets of most controlling tasks we care about are images synthesized from certain viewpoints.
The 2D rendering guidance can be implemented efficiently via replacing  $\vz_\theta(\vz_t)$ in \cref{eq:diff_sample} with  $\hat{\vz}_\theta(\vz_t)$ as:
\begin{equation}
    \hat{\vz}_\theta(\vz_t) = \vz_\theta(\vz_t) - w_t \nabla_{\vz_t}\ell\left[\vc, \mathcal{R}(\vz_\theta(\vz_t), \pi_c)\right], 
    \lbleq{guide}
\end{equation}
where $\vz_\theta$ is the denoised tri-plane derived from the unconditional prior, $w_t$ is the time-dependent weight.

\vspace{-5pt}\paragraph{Langevin correction steps} 
While the 2D rendering guidance can provide a gradient to learning 3D representations, the optimization is not often stable due to the nonlinearity of mapping from 2D to 3D. 
Our initial experiments showed that early guidance steps get stuck in a local minimum with incorrect geometry prediction, which is hard to correct in the later denoising stage when the noise level decreases. 
Therefore, we adopt similar ideas from the predictor-corrector~\cite{song2020score,ho2022video} to include additional Langevin correction steps before the diffusion step (\cref{eq:diff_sample}):
\begin{equation}
    \vz_t = \vz_t - \frac{1}{2}\delta\sigma_t\hat{\veps}_\theta(\vz_t) + \sqrt{\delta}\sigma_t\veps', \veps'\sim\mathcal{N}(0, I),
    \lbleq{langevin}
\end{equation}
where $\delta$ is the step size, and $\hat{\veps}_\theta$ is derived from $\hat{\vz}_\theta$ in \cref{eq:guide}. According to Langevin MCMC~\cite{ma2015complete}, the additional steps help $\vz_t$ match the marginal distribution given certain $\sigma_t$.

\paragraph{Discussion: Conditioning v.s. Guidance}
Compared to guidance methods in~\cref{sec:guidance}, training a conditional 3D diffusion model has several benefits. First, in guided diffusion, a proper-designed differentiable $\ell(.,.)$ is necessary to back-propagate the gradient guidance to the diffusion model, which, however, is not available for all kinds of conditional tasks. In contrast, conditional models do not have such requirements and can adapt any conditional distribution. Also, conditioning is computationally more efficient because the guidance requires rendering and back-propagating through the volume renderer $\mathcal{R}$ at each step.
However, conditioning methods have a possible issue. As we train our models based on the images generated by a pretrained 3D GAN~(\cref{eq:cond_diff}), the learned $p(\vz|\vc)$ probably has domain gaps between real images and synthesized images. In such a case, guidance-based methods become more reliable as $\ell$ is directly computed upon real controls.  

Optionally, we can combine the best of both worlds when $\ell$ is available. For instance, we learn a conditional diffusion model and generate samples jointly with guidance (see \cref{fig:pipeline} (c)). This paradigm can also be used when the test camera is not given: the guidance is used to update the camera $\pi_c$ predicted by the aforementioned conditional model. 

\section{Experimental Settings}
\setlength{\tabcolsep}{2pt}
\begin{table*}[tb!]
\begin{center}
\small
\caption{Quantitative comparison on inversion. 
Although optimizing the Tri-plane model can fit input views well, it falls short in generating realistic novel view images.
Overall, our method achieves the best performance. }
\label{table:inversion}
\begin{tabular}{llccccccc|ccccccc}
\toprule
&& \multicolumn{7}{c}{FFHQ}&\multicolumn{5}{|c}{AFHQ-Cat}\\
\midrule
&&PSNR $\uparrow$ &SSIM $\uparrow$ &LPIPS$\downarrow$  &ID$\uparrow$
&nvFID $\downarrow$ &nvKID $\uparrow$ &nvID$\uparrow$
&PSNR$\uparrow$ &SSIM $\uparrow$ &LPIPS$\downarrow$ & nvFID $\downarrow$  & nvKID $\uparrow$\\
\midrule
{\multirow{3}{*}{Opt.}}
&$\mathcal{W}$&15.93  &0.68 &0.42 &0.60 & 39.26  &0.023  &0.57 &16.08 &0.57 &0.42 &9.15&0.004\\
&$\mathcal{W+}$&17.91 &0.73 &0.34 &0.74 & 38.23  &0.022&0.68     &18.32 &0.62 &0.35&10.54&0.006\\
&\emph{Tri.}   &18.32 &0.78 &{\bf 0.11}&{\bf0.92} &138.0& 0.154&0.54  &17.53 &{\bf 0.71} &{\bf 0.14}&98.79&0.085\\
\multirow{1}{*}{Pred.}
&$\mathcal{W}$&14.82 &0.64 &0.54 &0.37 &45.06 &0.018&0.35 &14.56 &0.52 &0.55 &20.87 &0.006\\
\midrule
 \multicolumn{2}{c}{Ours} &{\bf 22.30}& {\bf 0.79} &0.23  &{0.89} & {\bf 13.48}&{\bf0.005}&{\bf0.81} 
&{\bf 20.11} & 0.66 & 0.24 &{\bf7.03}&{\bf0.003}
\\
\bottomrule
\vspace{-20pt}
\end{tabular}
\end{center}
\end{table*}








\paragraph{Dataset \& Tasks}
We evaluate {\model} on three standard image generation benchmarks -- FFHQ ($512^2$)~\cite{karras2019style}, AFHQ-cat ($512^2$)~\cite{Choi_2020_CVPR}, and ShapeNet ($128^2$)~\cite{Sitzmann2019} including two categories \textit{Cars} and \textit{Chairs}. 
Following EG3D~\cite{eg3d}, each image is associated with its camera pose.
\begin{figure}[t]
    \centering
    \includegraphics[width=\linewidth]{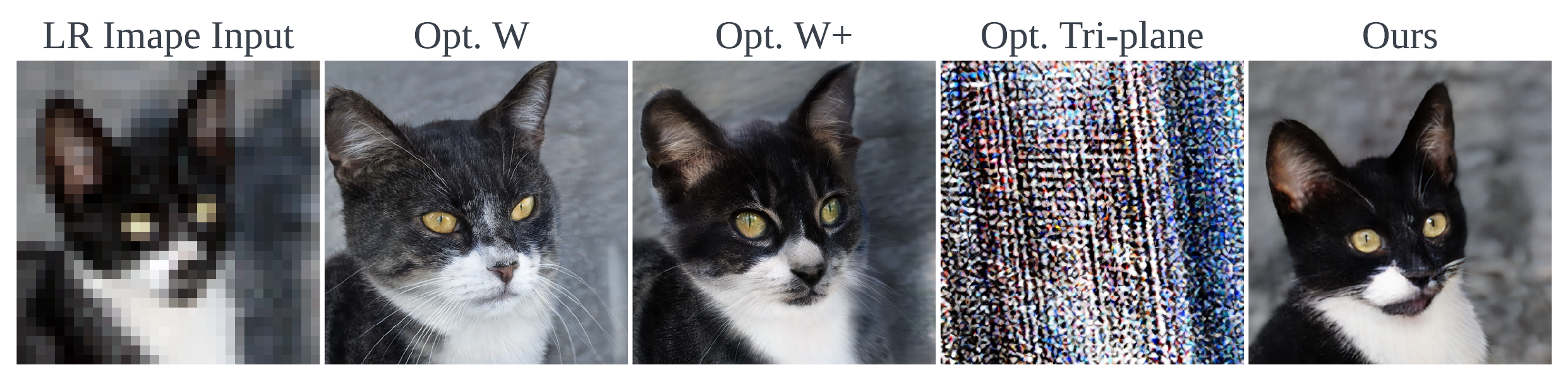}
    \caption{Comparison on the \textit{SR+inversion} task. By learning the proper prior with diffusion models, {\model} is able to synthesize realistic and faithful cat faces from low-resolution inputs, while optimization-based approaches fail completely due to the lack of proper 3D prior.}
    \vspace{-10pt}
    \lblfig{compare_sr}
\end{figure}

We consider \textit{six} controllable 3D-aware generation tasks. For all datasets, we test the standard \emph{image-to-3D inversion} (original resolution and low-resolution inputs) and \emph{edge-map to 3D} generations. For faces, we further explored \emph{segmentation to 3D}, \emph{head-shape to 3D} and \emph{text-description to 3D} tasks to validate the controllability at various levels. To compare with previous work~\cite{deng20233d} for \emph{Seg-to-3D}, we additionally train one model on CelebA-HQ~\cite{Karras2017}.
Besides, we also report performance on unconditional generation with guidance in the ablation.
\vspace{-10pt}\paragraph{Baselines}
We choose the standard optimization-based and encoder-based~\cite{tov2021e4e,ko20233d_gan_inversion} approaches for image-to-3D inversion, and the recent Pix2Pix3D~\cite{deng20233d} as the major baseline to compare on the \textit{Seg-to-3D} task.
Note that we do not focus on achieving the state-of-the-art on a single task like inversion, but rather to highlight the potential of our generic framework in 3D-aware generation. Thus, our comparison limits to methods without fine-tuning the model weights~\cite{roich2022pivotal}.

\vspace{-10pt}\paragraph{Evaluation Metrics}
For image synthesis quality, we report five standard metrics: PSNR, SSIM, SG diversity~\cite{chen2022sofgan},  LPIPS~\cite{Zhang2018f}, KID, and FID~\cite{fid}. 
For face, we compute the cosine similarity of the facial embeddings generated by the facial recognition network for a given pair of faces, utilizing it as ID metric.
In the context of conditional generation tasks, following Pix2Pix3D~\cite{deng20233d}, we evaluate methods using mean Intersection-over-Union (mIoU) and mean pixel accuracy (MPA) for segmentation maps.

\begin{table}[tb!]
\small
\begin{center}
\caption{Quantitative comparison on Seg2Face and Seg2Cat.}
\label{table:seg_edge}
\begin{tabular}{c|cccc|cccc}
\toprule
task& \multicolumn{4}{c|}{Seg2Face}&\multicolumn{4}{c}{Seg2Cat}\\
\midrule
metric&FID$\downarrow$&SG$\uparrow$&mIoU$\uparrow$&MPA$\uparrow$&FID$\downarrow$&SG$\uparrow$&mIoU$\uparrow$&MPA$\uparrow$\\
\midrule
p2p3D      &21.28 &{\bf0.46} &0.52 &0.63  &15.46  &{\bf0.50} &0.64 &0.76\\
ours       &{\bf 12.85} &0.43 &{\bf 0.61} & {\bf 0.72}  &{\bf 11.66}  &0.47 &{\bf0.67}&{\bf0.79}\\
\bottomrule
\end{tabular}
\end{center}
\end{table}


\vspace{-10pt}\paragraph{Implementation Details}
We implemented all our models based on the standard U-Net architectures~\cite{dhariwal2021diffusion} where for conditional diffusion models, an U-Net-based encoder is adopted to encode the input image similar to~\cite{gu2023nerfdiff}, see \cref{fig:pipeline} (b).
We include the hyper-parameter details in Appendix.

\section{Results}
We present both quantitative and qualitative results across image/seg/edge/text-to-3D synthesis, with a focus on FFHQ and AFHQ. Additional results and analysis on ShapeNet can be found in the \cref{sec:appendix_results}.
\subsection{Image-to-3D Inversion}
In this section, we evaluate {\model} on 3D inversion tasks, comparing our methods in two cases: (1) standard inversion and (2) a more challenging 3D super-resolution task. To establish a baseline, we directly optimize the low-dimensional latent vectors ($\mathcal{W}$, $\mathcal{W+}$),
following \cite{abdal2020image2stylegan++}, as well as triplanes. As conventional GANs do not have learned priors in these spaces, optimization is performed with noise injection regularization. We also employ an encoder-based approach \cite{ko20233d_gan_inversion} that directly predicts $\mathcal{W}$ or triplanes. To predict triplanes, we train a separate encoder.

The results are shown in \cref{table:inversion} and \cref{fig:compare_invert} where our methods significantly outperform the other methods in terms of both image quality and identity consistency. While direct optimization of triplanes may yield higher accuracy in the input view, it always results in collapsed novel view results due to a lack of prior. We also show the visual comparisons for 3D super-resolution in \cref{fig:compare_sr} where our diffusion-based approaches show more gains.

\begin{figure}[t]
    \centering
    \includegraphics[width=\linewidth]{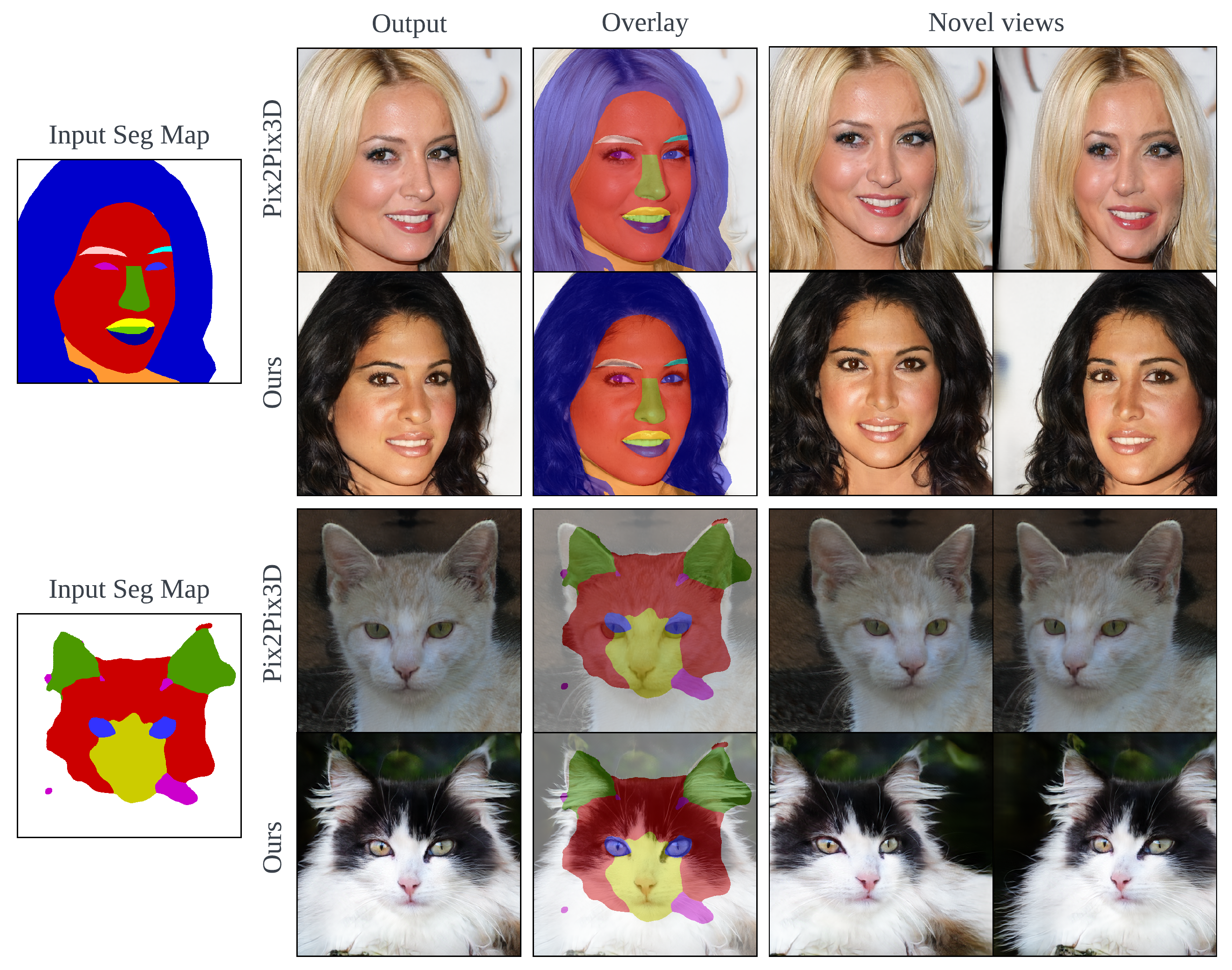}
    \caption{Comparison on \textit{Seg-to-3D} generation. {\bf All faces are model generated, and are not real identities.} Our proposed method generates images that achieve improved alignment with the segmentation map and greater 3D consistency.}
    \vspace{-10pt}
    \lblfig{compare_segmentation}
\end{figure}
\begin{figure}[t]
    \centering
    \includegraphics[width=\linewidth]{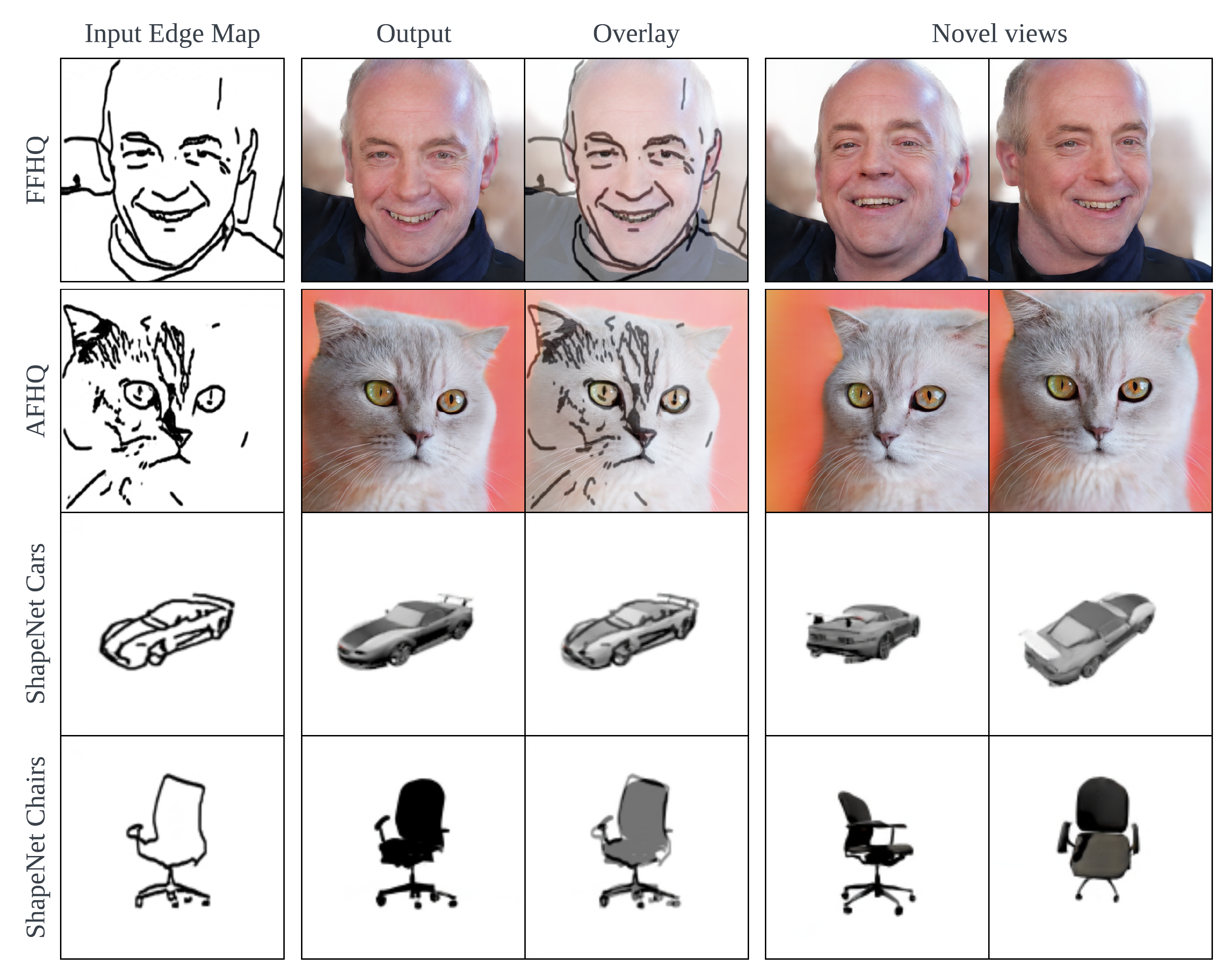}
    \caption{Qualitative results on \textit{Edge-to-3D} generation on all three datasets. {\bf All faces are model generated.} }
    \vspace{-10pt}
    \lblfig{compare_edge}
\end{figure}
\begin{figure*}[t]
    \centering
    \includegraphics[width=\linewidth]{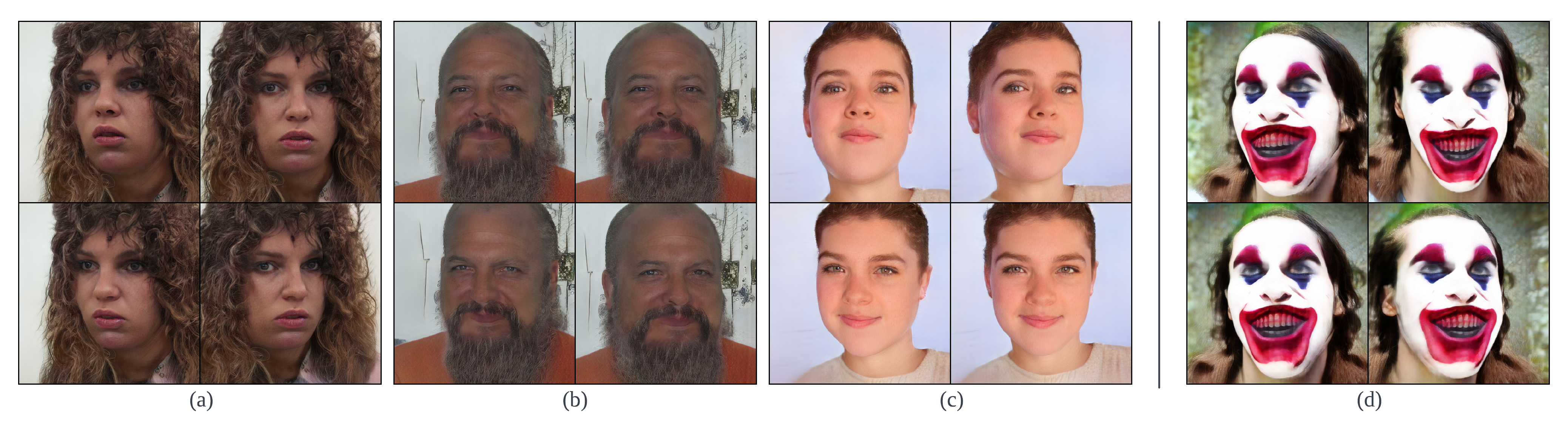}
    \vspace{-20pt}
    \caption{Qualitative results on \textit{Text-to-3D} synthesis based on given prompts: \textit{(a) A middle-aged woman with curly brown hair and pink lips; (b) A middle-aged man with a receding hairline, a thick beard, and hazel eyes; (c) A young woman with freckles on her cheeks and brown hair in a pixie cut; (d) a photography of Joker's face.} 
    \textbf{All faces are model generated, and are not real identities.} }
    \vspace{-10pt}
    \lblfig{compare_text}
\end{figure*}

\subsection{Seg-to-3D \& Edge-to-3D Synthesis}
We evaluate our methods on more general conditional 3D generation tasks where the input control is not necessarily the target view, e.g., \emph{Seg-to-3D} and \emph{Edge-to-3D} tasks. For the \emph{Seg-to-3D} task, we train two additional parsing networks \cite{yu2021bisenet} with labels provided by Pix2Pix3D \cite{deng20233d}, where the segmentation ground-truth of cats is obtained via clustering the DINO feature as proposed by \cite{amir2021deep_cluster}. It has been observed that this clustering scheme adversely affects the performance of cat-parsing networks, resulting in lower accuracy than that achieved by face-parsing networks.

The results of our evaluation are presented in \cref{table:seg_edge}, which indicates that our method generates images with comparable alignment and quality. Furthermore, as illustrated in \cref{fig:compare_segmentation}, our method is capable of producing more realistic faces in novel views. Additionally, our model successfully generates consistent 3D objects by taking as input edge maps, as illustrated in \cref{fig:compare_segmentation}. 

\subsection{Text-to-3D Synthesis}
We demonstrate the versatility of our framework by applying it to text-to-3D generation. The qualitative results are shown in \cref{fig:compare_text}. For (a)-(c), we train {\model} as a conditional diffusion model where we adopt the normalized CLIP embedding of the model's rendering as conditioning. In test time, such a model can be seamlessly switched to text-control thanks to the multi-modal space of CLIP. We also conduct experiments with text feature guidance in (d), where we directly apply a pre-trained 2D diffusion model as a score function similar to DreamFusion \cite{poole2022dreamfusion}, and guide the generation of an unconditional 3D diffusion model.

\subsection{Ablation Study}

\begin{figure}[t]
    \centering
    \begin{subfigure}[b]{0.5\textwidth}
    \includegraphics[width=\linewidth]{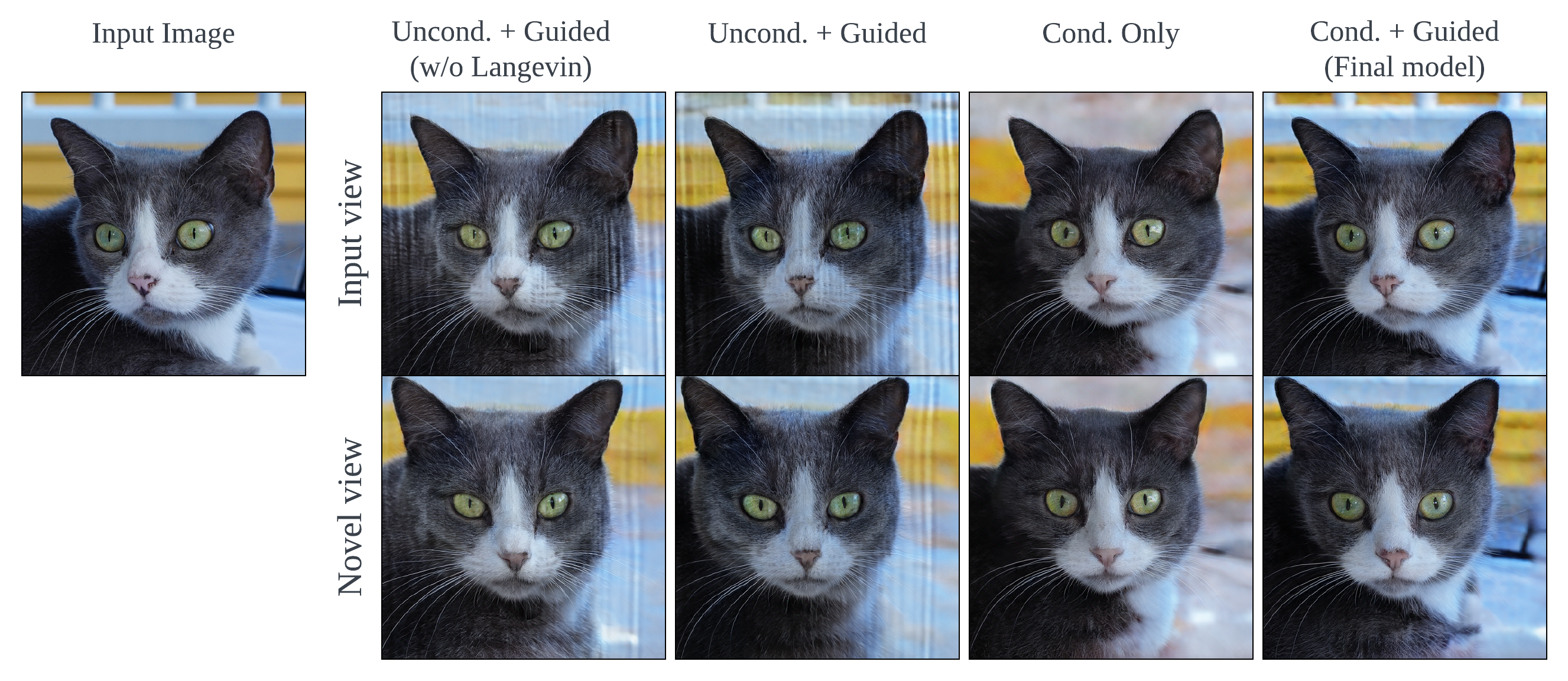}
    \caption{Visual comparison between conditional and guided diffusion.}
    \end{subfigure}
    \begin{subfigure}[b]{0.5\textwidth}
        \begin{center}
        \label{table:ablation}
        \begin{tabular}{l|ccccc}
        \toprule
        Metric & PSNR $\uparrow$&SSIM$\uparrow$&LPIPS$\downarrow$&nvFID$\downarrow$&nvKID$\downarrow$\\
        \midrule
        Cond. only & 14.52 & 0.50 & 0.52 & 8.13 & 0.003  \\
        Guided only & 16.75 & 0.59 & 0.35 & 8.52 & 0.004 \\
        Full (ours) & {\bf 20.11} &{\bf 0.66} & {\bf 0.24} & {\bf 7.03} & {\bf 0.003} \\
        \bottomrule
        \end{tabular}
        \caption{\label{tab:ablation} Numerical comparisons on generation strategies on AFHQ}
        \end{center}
    \end{subfigure}
    \caption{Ablation study on generation strategies.}
    \vspace{-15pt}
    \lblfig{compare_ablation}
\end{figure}

We conducted an ablation study on the task of Image-to-3D inversion, evaluating the effects of conditioning and guidance on the visual performance of our method. As illustrated in \cref{fig:compare_ablation}, our results demonstrate that the inclusion of conditioning and guidance leads to superior visual performance, while their absence results in artifacts and an inability to fit target images. More specifically, if we only apply guidance on unconditional models, the generated outputs seem to have artifacts. On the other hand, when using the conditional model only, the model is unable to recover all details from the input image especially for background. We also show a numerical comparison in~\cref{tab:ablation} to highlight the necessity of applying both conditioning and guidance. 

\section{Related Work}
\paragraph{Diffusion for 3D-aware Generation}
There have been recent attempts~\cite{rodin,bautista2022gaudi,pointe,mueller2022diffrf,anciukevicius2022renderdiffusion,triplanediff,Anonymous} to extend diffusion models to 3D. The key challenge here is to obtain 3D ground truth for training. Most works tackle this challenge by reconstructing 3D ground truth from dense multi-view data. Instead, our method can be trained only on single-view data by using a 3D GAN to synthesize infinite ground-truth 3D data. Another line of work~\cite{poole2022dreamfusion,SJC,zhou2022sparsefusion,nerdi,gu2023nerfdiff} applies 2D diffusion priors to the sparse-view reconstruction or text-to-3D generation tasks. For example, NerfDiff~\cite{gu2023nerfdiff} applies a test-time optimization by distilling  2D diffusion priors into NeRF for single-view reconstruction. Different from NerfDiff~\cite{gu2023nerfdiff}, our focus is   3D-aware image synthesis controlled by various control signals, and we apply denoising directly in 3D. Furthermore, our method can be trained on single-view datasets without the need of multi-view data.


\vspace{-10pt}
\paragraph{Controllable Image Synthesis with GANs}
Conventional GANs~\cite{goodfellow2014generative,karras2019style,karras2020analyzing} can generate photo-realistic images from low-dimensional randomly sampled latent vectors, but have limited controllability.  
Follow-up works enable controllability by either adding conditioning input along with the sampled vectors as input (named "Conditional GAN")~\cite{isola2017image,park2019semantic} or manipulating the sampled vectors~\cite{shen2020closed,harkonen2020ganspace,zhu2023linkgan}. These works only focus on 2D image synthesis with control, which cannot explicitly control 3D properties (e.g., cameras) and synthesize multi-view consistent images. Recently, 3D-GANs~\cite{schwarz2020graf,eg3d,Chan2021Pi-GAN:Synthesis,Niemeyer2020GIRAFFE,gu2021stylenerf,xu2021volumegan} have been developed by integrating 3D representation and rendering into GANs. While these models can control 3D properties by manipulating the latent vectors, their controllability is limited to global camera poses or geometry. Many works~\cite{sun2022next,sun2022ide,bergman2022gnarf,jiang2022faceediting} support fine-grained geometry editing, but most of them have only demonstrated results on human face or body. Other conditional 3D GANs for general objects~\cite{pix2nerf,deng20233d} need additional constraints or architecture changes, however, their synthesis quality is still limited. In contrast, our method allows a variety of control signals (e.g., segmentation map) for fine-level 3D-aware image synthesis on various kinds of objects. 




\section{Conclusion}

In summary, we propose {\model}, a versatile approach for 3D-aware image synthesis that combines the strengths of 3D GANs and diffusion models. Our method enables precise control over image synthesis by explicitly modeling the underlying latent distribution. We validate our approach on standard benchmarks, demonstrating its efficacy with various types of conditioning inputs. {\model} represents a significant advancement in generative modeling in 3D, opening up new research possibilities in this area.

{
    \small
    \bibliographystyle{ieeenat_fullname}
    \bibliography{egbib}

\begin{thebibliography}{105}
\providecommand{\natexlab}[1]{#1}
\providecommand{\url}[1]{\texttt{#1}}
\expandafter\ifx\csname urlstyle\endcsname\relax
  \providecommand{\doi}[1]{doi: #1}\else
  \providecommand{\doi}{doi: \begingroup \urlstyle{rm}\Url}\fi

\bibitem[Abdal et~al.(2019)Abdal, Qin, and
  Wonka]{abdal2019image2stylegan_wplus}
Rameen Abdal, Yipeng Qin, and Peter Wonka.
\newblock Image2stylegan: How to embed images into the stylegan latent space?
\newblock In \emph{Proceedings of the IEEE/CVF International Conference on
  Computer Vision}, pages 4432--4441, 2019.

\bibitem[Abdal et~al.(2020)Abdal, Qin, and Wonka]{abdal2020image2stylegan++}
Rameen Abdal, Yipeng Qin, and Peter Wonka.
\newblock Image2stylegan++: How to edit the embedded images?
\newblock In \emph{Proceedings of the IEEE/CVF conference on computer vision
  and pattern recognition}, pages 8296--8305, 2020.

\bibitem[Amir et~al.(2021)Amir, Gandelsman, Bagon, and
  Dekel]{amir2021deep_cluster}
Shir Amir, Yossi Gandelsman, Shai Bagon, and Tali Dekel.
\newblock Deep vit features as dense visual descriptors.
\newblock \emph{arXiv preprint arXiv:2112.05814}, 2\penalty0 (3):\penalty0 4,
  2021.

\bibitem[Anciukevicius et~al.(2022)Anciukevicius, Xu, Fisher, Henderson, Bilen,
  Mitra, and Guerrero]{anciukevicius2022renderdiffusion}
Titas Anciukevicius, Zexiang Xu, Matthew Fisher, Paul Henderson, Hakan Bilen,
  Niloy~J. Mitra, and Paul Guerrero.
\newblock {RenderDiffusion}: Image diffusion for {3D} reconstruction,
  inpainting and generation.
\newblock \emph{arXiv}, 2022.

\bibitem[Anonymous(SEE COVER LETTER)]{Anonymous}
Anonymous.
\newblock Anonymous, SEE COVER LETTER.

\bibitem[Bansal et~al.(2023)Bansal, Chu, Schwarzschild, Sengupta, Goldblum,
  Geiping, and Goldstein]{bansal2023universal}
Arpit Bansal, Hong-Min Chu, Avi Schwarzschild, Soumyadip Sengupta, Micah
  Goldblum, Jonas Geiping, and Tom Goldstein.
\newblock Universal guidance for diffusion models.
\newblock \emph{arXiv preprint arXiv:2302.07121}, 2023.

\bibitem[Bautista et~al.(2022)Bautista, Guo, Abnar, Talbott, Toshev, Chen,
  Dinh, Zhai, Goh, Ulbricht, Dehghan, and Susskind]{bautista2022gaudi}
Miguel~Angel Bautista, Pengsheng Guo, Samira Abnar, Walter Talbott, Alexander
  Toshev, Zhuoyuan Chen, Laurent Dinh, Shuangfei Zhai, Hanlin Goh, Daniel
  Ulbricht, Afshin Dehghan, and Josh Susskind.
\newblock Gaudi: A neural architect for immersive 3d scene generation.
\newblock \emph{arXiv}, 2022.

\bibitem[Bergman et~al.(2022)Bergman, Kellnhofer, Yifan, Chan, Lindell, and
  Wetzstein]{bergman2022gnarf}
Alexander~W. Bergman, Petr Kellnhofer, Wang Yifan, Eric~R. Chan, David~B.
  Lindell, and Gordon Wetzstein.
\newblock Generative neural articulated radiance fields.
\newblock In \emph{NeurIPS}, 2022.

\bibitem[Cai et~al.(2022)Cai, Obukhov, Dai, and Van~Gool]{pix2nerf}
Shengqu Cai, Anton Obukhov, Dengxin Dai, and Luc Van~Gool.
\newblock Pix2nerf: Unsupervised conditional $\pi$-gan for single image to
  neural radiance fields translation.
\newblock \emph{arXiv preprint arXiv:2202.13162}, 2022.

\bibitem[Chan et~al.(2022)Chan, Durand, and
  Isola]{chan2022learning_drawingline}
Caroline Chan, Fr{\'e}do Durand, and Phillip Isola.
\newblock Learning to generate line drawings that convey geometry and
  semantics.
\newblock In \emph{Proceedings of the IEEE/CVF Conference on Computer Vision
  and Pattern Recognition}, pages 7915--7925, 2022.

\bibitem[Chan et~al.(2020)Chan, Monteiro, Kellnhofer, Wu, and
  Wetzstein]{chanmonteiro2020pi-GAN}
Eric Chan, Marco Monteiro, Petr Kellnhofer, Jiajun Wu, and Gordon Wetzstein.
\newblock pi-gan: Periodic implicit generative adversarial networks for
  3d-aware image synthesis.
\newblock In \emph{arXiv}, 2020.

\bibitem[Chan et~al.(2021{\natexlab{a}})Chan, Lin, Chan, Nagano, Pan, De~Mello,
  Gallo, Guibas, Tremblay, Khamis, et~al.]{eg3d}
Eric~R Chan, Connor~Z Lin, Matthew~A Chan, Koki Nagano, Boxiao Pan, Shalini
  De~Mello, Orazio Gallo, Leonidas Guibas, Jonathan Tremblay, Sameh Khamis,
  et~al.
\newblock Efficient geometry-aware 3d generative adversarial networks.
\newblock \emph{arXiv preprint arXiv:2112.07945}, 2021{\natexlab{a}}.

\bibitem[Chan et~al.(2021{\natexlab{b}})Chan, Monteiro, Kellnhofer, Wu, and
  Wetzstein]{Chan2021Pi-GAN:Synthesis}
Eric~R. Chan, Marco Monteiro, Petr Kellnhofer, Jiajun Wu, and Gordon Wetzstein.
\newblock {pi-GAN: Periodic Implicit Generative Adversarial Networks for
  3D-Aware Image Synthesis}.
\newblock \emph{CVPR}, 2021{\natexlab{b}}.

\bibitem[Chen et~al.(2022{\natexlab{a}})Chen, Liu, Xie, Chen, Su, and
  Yu]{chen2022sofgan}
Anpei Chen, Ruiyang Liu, Ling Xie, Zhang Chen, Hao Su, and Jingyi Yu.
\newblock Sofgan: A portrait image generator with dynamic styling.
\newblock \emph{ACM Transactions on Graphics (TOG)}, 41\penalty0 (1):\penalty0
  1--26, 2022{\natexlab{a}}.

\bibitem[Chen et~al.(2022{\natexlab{b}})Chen, Xu, Geiger, Yu, and Su]{tensorf}
Anpei Chen, Zexiang Xu, Andreas Geiger, Jingyi Yu, and Hao Su.
\newblock Tensorf: Tensorial radiance fields.
\newblock \emph{arXiv preprint arXiv:2203.09517}, 2022{\natexlab{b}}.

\bibitem[Choi et~al.(2020)Choi, Uh, Yoo, and Ha]{Choi_2020_CVPR}
Yunjey Choi, Youngjung Uh, Jaejun Yoo, and Jung-Woo Ha.
\newblock Stargan v2: Diverse image synthesis for multiple domains.
\newblock In \emph{Proceedings of the IEEE/CVF Conference on Computer Vision
  and Pattern Recognition (CVPR)}, 2020.

\bibitem[Chung et~al.(2022)Chung, Kim, Mccann, Klasky, and
  Ye]{chung2022diffusion}
Hyungjin Chung, Jeongsol Kim, Michael~T Mccann, Marc~L Klasky, and Jong~Chul
  Ye.
\newblock Diffusion posterior sampling for general noisy inverse problems.
\newblock \emph{arXiv preprint arXiv:2209.14687}, 2022.

\bibitem[Deng et~al.(2022)Deng, Jiang, Qi, Yan, Zhou, Guibas, and
  Anguelov]{nerdi}
Congyue Deng, Chiyu~"Max'' Jiang, Charles~R. Qi, Xinchen Yan, Yin Zhou,
  Leonidas Guibas, and Dragomir Anguelov.
\newblock Nerdi: Single-view nerf synthesis with language-guided diffusion as
  general image priors, 2022.

\bibitem[Deng et~al.(2023)Deng, Yang, Ramanan, and Zhu]{deng20233d}
Kangle Deng, Gengshan Yang, Deva Ramanan, and Jun-Yan Zhu.
\newblock 3d-aware conditional image synthesis.
\newblock \emph{arXiv preprint arXiv:2302.08509}, 2023.

\bibitem[Dhariwal and Nichol(2021)]{dhariwal2021diffusion}
Prafulla Dhariwal and Alexander Nichol.
\newblock Diffusion models beat gans on image synthesis.
\newblock \emph{Advances in Neural Information Processing Systems},
  34:\penalty0 8780--8794, 2021.

\bibitem[Feng et~al.(2021)Feng, Feng, Black, and
  Bolkart]{feng2021learning_deca}
Yao Feng, Haiwen Feng, Michael~J Black, and Timo Bolkart.
\newblock Learning an animatable detailed 3d face model from in-the-wild
  images.
\newblock \emph{ACM Transactions on Graphics (ToG)}, 40\penalty0 (4):\penalty0
  1--13, 2021.

\bibitem[Goodfellow et~al.(2014{\natexlab{a}})Goodfellow, Pouget-Abadie, Mirza,
  Xu, Warde-Farley, Ozair, Courville, and Bengio]{goodfellow2014generative}
Ian Goodfellow, Jean Pouget-Abadie, Mehdi Mirza, Bing Xu, David Warde-Farley,
  Sherjil Ozair, Aaron Courville, and Yoshua Bengio.
\newblock Generative adversarial nets.
\newblock In \emph{NeurIPS}, 2014{\natexlab{a}}.

\bibitem[Goodfellow et~al.(2014{\natexlab{b}})Goodfellow, Pouget-Abadie, Mirza,
  Xu, Warde-Farley, Ozair, Courville, and Bengio]{Goodfellow2014}
Ian~J. Goodfellow, Jean Pouget-Abadie, Mehdi Mirza, Bing Xu, David
  Warde-Farley, Sherjil Ozair, Aaron Courville, and Yoshua Bengio.
\newblock {Generative Adversarial Nets}.
\newblock \emph{Advances in Neural Information Processing Systems}, pages
  2672--2680, 2014{\natexlab{b}}.

\bibitem[Graikos et~al.(2022)Graikos, Malkin, Jojic, and
  Samaras]{graikos2022diffusion}
Alexandros Graikos, Nikolay Malkin, Nebojsa Jojic, and Dimitris Samaras.
\newblock Diffusion models as plug-and-play priors.
\newblock \emph{arXiv preprint arXiv:2206.09012}, 2022.

\bibitem[Gu et~al.(2021)Gu, Liu, Wang, and Theobalt]{gu2021stylenerf}
Jiatao Gu, Lingjie Liu, Peng Wang, and Christian Theobalt.
\newblock Stylenerf: A style-based 3d-aware generator for high-resolution image
  synthesis.
\newblock \emph{arXiv preprint arXiv:2110.08985}, 2021.

\bibitem[Gu et~al.(2022)Gu, Zhai, Zhang, Bautista, and Susskind]{gu2022f}
Jiatao Gu, Shuangfei Zhai, Yizhe Zhang, Miguel~Angel Bautista, and Josh
  Susskind.
\newblock f-dm: A multi-stage diffusion model via progressive signal
  transformation.
\newblock \emph{arXiv preprint arXiv:2210.04955}, 2022.

\bibitem[Gu et~al.(2023)Gu, Trevithick, Lin, Susskind, Theobalt, Liu, and
  Ramamoorthi]{gu2023nerfdiff}
Jiatao Gu, Alex Trevithick, Kai-En Lin, Josh Susskind, Christian Theobalt,
  Lingjie Liu, and Ravi Ramamoorthi.
\newblock Nerfdiff: Single-image view synthesis with nerf-guided distillation
  from 3d-aware diffusion.
\newblock \emph{arXiv preprint arXiv:2302.10109}, 2023.

\bibitem[H{\"a}rk{\"o}nen et~al.(2020)H{\"a}rk{\"o}nen, Hertzmann, Lehtinen,
  and Paris]{harkonen2020ganspace}
Erik H{\"a}rk{\"o}nen, Aaron Hertzmann, Jaakko Lehtinen, and Sylvain Paris.
\newblock Ganspace: Discovering interpretable gan controls.
\newblock \emph{arXiv preprint arXiv:2004.02546}, 2020.

\bibitem[Heusel et~al.(2017)Heusel, Ramsauer, Unterthiner, Nessler, and
  Hochreiter]{fid}
Martin Heusel, Hubert Ramsauer, Thomas Unterthiner, Bernhard Nessler, and Sepp
  Hochreiter.
\newblock Gans trained by a two time-scale update rule converge to a local nash
  equilibrium.
\newblock \emph{Advances in neural information processing systems}, 30, 2017.

\bibitem[Ho and Salimans(2022)]{ho2022classifier}
Jonathan Ho and Tim Salimans.
\newblock Classifier-free diffusion guidance.
\newblock \emph{arXiv preprint arXiv:2207.12598}, 2022.

\bibitem[Ho et~al.(2020)Ho, Jain, and Abbeel]{ho2020denoising}
Jonathan Ho, Ajay Jain, and Pieter Abbeel.
\newblock Denoising diffusion probabilistic models.
\newblock \emph{Advances in Neural Information Processing Systems},
  33:\penalty0 6840--6851, 2020.

\bibitem[Ho et~al.(2022{\natexlab{a}})Ho, Chan, Saharia, Whang, Gao, Gritsenko,
  Kingma, Poole, Norouzi, Fleet, et~al.]{ho2022imagen}
Jonathan Ho, William Chan, Chitwan Saharia, Jay Whang, Ruiqi Gao, Alexey
  Gritsenko, Diederik~P Kingma, Ben Poole, Mohammad Norouzi, David~J Fleet,
  et~al.
\newblock Imagen video: High definition video generation with diffusion models.
\newblock \emph{arXiv preprint arXiv:2210.02303}, 2022{\natexlab{a}}.

\bibitem[Ho et~al.(2022{\natexlab{b}})Ho, Salimans, Gritsenko, Chan, Norouzi,
  and Fleet]{ho2022video}
Jonathan Ho, Tim Salimans, Alexey Gritsenko, William Chan, Mohammad Norouzi,
  and David~J Fleet.
\newblock Video diffusion models.
\newblock \emph{arXiv preprint arXiv:2204.03458}, 2022{\natexlab{b}}.

\bibitem[Hoogeboom et~al.(2023)Hoogeboom, Heek, and
  Salimans]{hoogeboom2023simple}
Emiel Hoogeboom, Jonathan Heek, and Tim Salimans.
\newblock simple diffusion: End-to-end diffusion for high resolution images.
\newblock \emph{arXiv preprint arXiv:2301.11093}, 2023.

\bibitem[Isola et~al.(2017)Isola, Zhu, Zhou, and Efros]{isola2017image}
Phillip Isola, Jun-Yan Zhu, Tinghui Zhou, and Alexei~A Efros.
\newblock Image-to-image translation with conditional adversarial networks.
\newblock In \emph{CVPR}, 2017.

\bibitem[Jiang et~al.(2022)Jiang, Chen, Liu, Fu, and Gao]{jiang2022faceediting}
Kaiwen Jiang, Shu-Yu Chen, Feng-Lin Liu, Hongbo Fu, and Lin Gao.
\newblock Nerffaceediting: Disentangled face editing in neural radiance fields,
  2022.

\bibitem[Jingxiang et~al.(2022)Jingxiang, Xuan, Lizhen, Xiaoyu, Yong, Hongwen,
  and Yebin]{sun2022next}
Sun Jingxiang, Wang Xuan, Wang Lizhen, Li Xiaoyu, Zhang Yong, Zhang Hongwen,
  and Liu Yebin.
\newblock Next3d: Generative neural texture rasterization for 3d-aware head
  avatars.
\newblock \emph{arXiv preprint arXiv:2205.15517}, 2022.

\bibitem[Karras et~al.(2018)Karras, Aila, Laine, and Lehtinen]{Karras2017}
Tero Karras, Timo Aila, Samuli Laine, and Jaakko Lehtinen.
\newblock {Progressive Growing of GANs for Improved Quality, Stability, and
  Variation}.
\newblock \emph{International Conference on Learning Representations}, 2018.

\bibitem[Karras et~al.(2019)Karras, Laine, and Aila]{karras2019style}
Tero Karras, Samuli Laine, and Timo Aila.
\newblock A style-based generator architecture for generative adversarial
  networks.
\newblock In \emph{CVPR}, pages 4401--4410, 2019.

\bibitem[Karras et~al.(2020{\natexlab{a}})Karras, Aittala, Hellsten, Laine,
  Lehtinen, and Aila]{Karras2020ada}
Tero Karras, Miika Aittala, Janne Hellsten, Samuli Laine, Jaakko Lehtinen, and
  Timo Aila.
\newblock Training generative adversarial networks with limited data.
\newblock In \emph{Proc. NeurIPS}, 2020{\natexlab{a}}.

\bibitem[Karras et~al.(2020{\natexlab{b}})Karras, Laine, Aittala, Hellsten,
  Lehtinen, and Aila]{karras2020analyzing}
Tero Karras, Samuli Laine, Miika Aittala, Janne Hellsten, Jaakko Lehtinen, and
  Timo Aila.
\newblock Analyzing and improving the image quality of stylegan.
\newblock In \emph{CVPR}, pages 8110--8119, 2020{\natexlab{b}}.

\bibitem[Karras et~al.(2020{\natexlab{c}})Karras, Laine, Aittala, Hellsten,
  Lehtinen, and Aila]{karras2020analyzing_inversion}
Tero Karras, Samuli Laine, Miika Aittala, Janne Hellsten, Jaakko Lehtinen, and
  Timo Aila.
\newblock Analyzing and improving the image quality of stylegan.
\newblock In \emph{Proceedings of the IEEE/CVF conference on computer vision
  and pattern recognition}, pages 8110--8119, 2020{\natexlab{c}}.

\bibitem[Karras et~al.(2021)Karras, Aittala, Laine, H\"ark\"onen, Hellsten,
  Lehtinen, and Aila]{karras2021}
Tero Karras, Miika Aittala, Samuli Laine, Erik H\"ark\"onen, Janne Hellsten,
  Jaakko Lehtinen, and Timo Aila.
\newblock Alias-free generative adversarial networks.
\newblock In \emph{Proc. NeurIPS}, 2021.

\bibitem[Kawar et~al.(2022)Kawar, Elad, Ermon, and Song]{kawar2022denoising}
Bahjat Kawar, Michael Elad, Stefano Ermon, and Jiaming Song.
\newblock Denoising diffusion restoration models.
\newblock \emph{arXiv preprint arXiv:2201.11793}, 2022.

\bibitem[Kingma and Ba(2014)]{kingma2014adam}
Diederik~P Kingma and Jimmy Ba.
\newblock Adam: A method for stochastic optimization.
\newblock \emph{arXiv preprint arXiv:1412.6980}, 2014.

\bibitem[Ko et~al.(2023{\natexlab{a}})Ko, Cho, Choi, Ryoo, and Kim]{ko20233d}
Jaehoon Ko, Kyusun Cho, Daewon Choi, Kwangrok Ryoo, and Seungryong Kim.
\newblock 3d gan inversion with pose optimization.
\newblock In \emph{Proceedings of the IEEE/CVF Winter Conference on
  Applications of Computer Vision}, pages 2967--2976, 2023{\natexlab{a}}.

\bibitem[Ko et~al.(2023{\natexlab{b}})Ko, Cho, Choi, Ryoo, and
  Kim]{ko20233d_gan_inversion}
Jaehoon Ko, Kyusun Cho, Daewon Choi, Kwangrok Ryoo, and Seungryong Kim.
\newblock 3d gan inversion with pose optimization.
\newblock In \emph{Proceedings of the IEEE/CVF Winter Conference on
  Applications of Computer Vision}, pages 2967--2976, 2023{\natexlab{b}}.

\bibitem[Kosiorek et~al.(2021)Kosiorek, Strathmann, Zoran, Moreno, Schneider,
  Mokr{\'{a}}, and Rezende]{nerfvae}
Adam~R. Kosiorek, Heiko Strathmann, Daniel Zoran, Pol Moreno, Rosalia
  Schneider, Soňa Mokr{\'{a}}, and Danilo~J. Rezende.
\newblock {NeRF-VAE: A Geometry Aware 3D Scene Generative Model}.
\newblock \emph{ICML}, 2021.

\bibitem[Li et~al.(2022{\natexlab{a}})Li, Yang, Chang, Chen, Feng, Xu, Li, and
  Chen]{li2022srdiff}
Haoying Li, Yifan Yang, Meng Chang, Shiqi Chen, Huajun Feng, Zhihai Xu, Qi Li,
  and Yueting Chen.
\newblock Srdiff: Single image super-resolution with diffusion probabilistic
  models.
\newblock \emph{Neurocomputing}, 479:\penalty0 47--59, 2022{\natexlab{a}}.

\bibitem[Li et~al.(2017)Li, Bolkart, Black, Li, and
  Romero]{li2017learning_flame}
Tianye Li, Timo Bolkart, Michael~J Black, Hao Li, and Javier Romero.
\newblock Learning a model of facial shape and expression from 4d scans.
\newblock \emph{ACM Trans. Graph.}, 36\penalty0 (6):\penalty0 194--1, 2017.

\bibitem[Li et~al.(2022{\natexlab{b}})Li, Xu, Wu, Zheng, Dai, Pumarola, Zhang,
  Vajda, and Kitani]{li20223d}
Yu-Jhe Li, Tao Xu, Bichen Wu, Ningyuan Zheng, Xiaoliang Dai, Albert Pumarola,
  Peizhao Zhang, Peter Vajda, and Kris Kitani.
\newblock 3d-aware encoding for style-based neural radiance fields.
\newblock \emph{arXiv preprint arXiv:2211.06583}, 2022{\natexlab{b}}.

\bibitem[Lin et~al.(2022)Lin, Lindell, Chan, and Wetzstein]{lin20223d}
Connor~Z Lin, David~B Lindell, Eric~R Chan, and Gordon Wetzstein.
\newblock 3d gan inversion for controllable portrait image animation.
\newblock \emph{arXiv preprint arXiv:2203.13441}, 2022.

\bibitem[Loshchilov and Hutter(2017)]{loshchilov2017decoupled}
Ilya Loshchilov and Frank Hutter.
\newblock Decoupled weight decay regularization.
\newblock \emph{arXiv preprint arXiv:1711.05101}, 2017.

\bibitem[Lugmayr et~al.(2022)Lugmayr, Danelljan, Romero, Yu, Timofte, and
  Van~Gool]{lugmayr2022repaint}
Andreas Lugmayr, Martin Danelljan, Andres Romero, Fisher Yu, Radu Timofte, and
  Luc Van~Gool.
\newblock Repaint: Inpainting using denoising diffusion probabilistic models.
\newblock In \emph{Proceedings of the IEEE/CVF Conference on Computer Vision
  and Pattern Recognition}, pages 11461--11471, 2022.

\bibitem[Ma et~al.(2015)Ma, Chen, and Fox]{ma2015complete}
Yi-An Ma, Tianqi Chen, and Emily Fox.
\newblock A complete recipe for stochastic gradient mcmc.
\newblock \emph{Advances in neural information processing systems}, 28, 2015.

\bibitem[Max(1995)]{max}
Nelson Max.
\newblock Optical models for direct volume rendering.
\newblock \emph{IEEE Transactions on Visualization and Computer Graphics},
  1\penalty0 (2):\penalty0 99--108, 1995.

\bibitem[Mildenhall et~al.(2020)Mildenhall, Srinivasan, Tancik, Barron,
  Ramamoorthi, and Ng]{mildenhall2020nerf}
Ben Mildenhall, Pratul~P Srinivasan, Matthew Tancik, Jonathan~T Barron, Ravi
  Ramamoorthi, and Ren Ng.
\newblock Nerf: Representing scenes as neural radiance fields for view
  synthesis.
\newblock In \emph{European conference on computer vision}, pages 405--421.
  Springer, 2020.

\bibitem[Mohamed and Lakshminarayanan(2016)]{mohamed2016learning}
Shakir Mohamed and Balaji Lakshminarayanan.
\newblock Learning in implicit generative models.
\newblock \emph{arXiv preprint arXiv:1610.03483}, 2016.

\bibitem[M{\"{u}}ller et~al.()M{\"{u}}ller, , Siddiqui, Porzi, Rota~Bul\`{o},
  Kontschieder, and Nie{\ss}ner]{mueller2022diffrf}
Norman M{\"{u}}ller, , Yawar Siddiqui, Lorenzo Porzi, Samuel Rota~Bul\`{o},
  Peter Kontschieder, and Matthias Nie{\ss}ner.

\bibitem[M{\"u}ller et~al.(2022)M{\"u}ller, Siddiqui, Porzi, Bul{\`o},
  Kontschieder, and Nie{\ss}ner]{muller2022diffrf}
Norman M{\"u}ller, Yawar Siddiqui, Lorenzo Porzi, Samuel~Rota Bul{\`o}, Peter
  Kontschieder, and Matthias Nie{\ss}ner.
\newblock Diffrf: Rendering-guided 3d radiance field diffusion.
\newblock \emph{arXiv preprint arXiv:2212.01206}, 2022.

\bibitem[Nichol et~al.(2022)Nichol, Jun, Dhariwal, Mishkin, and Chen]{pointe}
Alex Nichol, Heewoo Jun, Prafulla Dhariwal, Pamela Mishkin, and Mark Chen.
\newblock Point-e: A system for generating 3d point clouds from complex
  prompts, 2022.

\bibitem[Niemeyer and Geiger(2021{\natexlab{a}})]{Niemeyer2020GIRAFFE}
Michael Niemeyer and Andreas Geiger.
\newblock Giraffe: Representing scenes as compositional generative neural
  feature fields.
\newblock In \emph{Proc. IEEE Conf. on Computer Vision and Pattern Recognition
  (CVPR)}, 2021{\natexlab{a}}.

\bibitem[Niemeyer and Geiger(2021{\natexlab{b}})]{giraffe}
Michael Niemeyer and Andreas Geiger.
\newblock {GIRAFFE: Representing Scenes as Compositional Generative Neural
  Feature Fields}.
\newblock \emph{CVPR}, pages 11453--11464, 2021{\natexlab{b}}.

\bibitem[Or-El et~al.(2022)Or-El, Luo, Shan, Shechtman, Park, and
  Kemelmacher-Shlizerman]{or2022stylesdf}
Roy Or-El, Xuan Luo, Mengyi Shan, Eli Shechtman, Jeong~Joon Park, and Ira
  Kemelmacher-Shlizerman.
\newblock Stylesdf: High-resolution 3d-consistent image and geometry
  generation.
\newblock In \emph{Proceedings of the IEEE/CVF Conference on Computer Vision
  and Pattern Recognition}, pages 13503--13513, 2022.

\bibitem[Park et~al.(2019)Park, Liu, Wang, and Zhu]{park2019semantic}
Taesung Park, Ming-Yu Liu, Ting-Chun Wang, and Jun-Yan Zhu.
\newblock Semantic image synthesis with spatially-adaptive normalization.
\newblock In \emph{CVPR}, 2019.

\bibitem[Peebles and Xie(2022)]{peebles2022scalable}
William Peebles and Saining Xie.
\newblock Scalable diffusion models with transformers.
\newblock \emph{arXiv preprint arXiv:2212.09748}, 2022.

\bibitem[Poole et~al.(2022)Poole, Jain, Barron, and
  Mildenhall]{poole2022dreamfusion}
Ben Poole, Ajay Jain, Jonathan~T Barron, and Ben Mildenhall.
\newblock Dreamfusion: Text-to-3d using 2d diffusion.
\newblock \emph{arXiv preprint arXiv:2209.14988}, 2022.

\bibitem[Radford et~al.(2021{\natexlab{a}})Radford, Kim, Hallacy, Ramesh, Goh,
  Agarwal, Sastry, Askell, Mishkin, Clark, et~al.]{radford2021learning}
Alec Radford, Jong~Wook Kim, Chris Hallacy, Aditya Ramesh, Gabriel Goh,
  Sandhini Agarwal, Girish Sastry, Amanda Askell, Pamela Mishkin, Jack Clark,
  et~al.
\newblock Learning transferable visual models from natural language
  supervision.
\newblock In \emph{International conference on machine learning}, pages
  8748--8763. PMLR, 2021{\natexlab{a}}.

\bibitem[Radford et~al.(2021{\natexlab{b}})Radford, Kim, Hallacy, Ramesh, Goh,
  Agarwal, Sastry, Askell, Mishkin, Clark, et~al.]{radford2021learning_clip}
Alec Radford, Jong~Wook Kim, Chris Hallacy, Aditya Ramesh, Gabriel Goh,
  Sandhini Agarwal, Girish Sastry, Amanda Askell, Pamela Mishkin, Jack Clark,
  et~al.
\newblock Learning transferable visual models from natural language
  supervision.
\newblock In \emph{International conference on machine learning}, pages
  8748--8763. PMLR, 2021{\natexlab{b}}.

\bibitem[Ramesh et~al.(2022)Ramesh, Dhariwal, Nichol, Chu, and
  Chen]{ramesh2022hierarchical}
Aditya Ramesh, Prafulla Dhariwal, Alex Nichol, Casey Chu, and Mark Chen.
\newblock Hierarchical text-conditional image generation with clip latents.
\newblock \emph{arXiv preprint arXiv:2204.06125}, 2022.

\bibitem[Rebain et~al.(2022)Rebain, Matthews, Yi, Lagun, and
  Tagliasacchi]{rebain2022lolnerf}
Daniel Rebain, Mark Matthews, Kwang~Moo Yi, Dmitry Lagun, and Andrea
  Tagliasacchi.
\newblock Lolnerf: Learn from one look.
\newblock In \emph{Proceedings of the IEEE/CVF Conference on Computer Vision
  and Pattern Recognition}, pages 1558--1567, 2022.

\bibitem[Roich et~al.(2022)Roich, Mokady, Bermano, and
  Cohen-Or]{roich2022pivotal}
Daniel Roich, Ron Mokady, Amit~H Bermano, and Daniel Cohen-Or.
\newblock Pivotal tuning for latent-based editing of real images.
\newblock \emph{ACM Transactions on Graphics (TOG)}, 42\penalty0 (1):\penalty0
  1--13, 2022.

\bibitem[Rombach et~al.(2021)Rombach, Blattmann, Lorenz, Esser, and
  Ommer]{rombach2021highresolution}
Robin Rombach, Andreas Blattmann, Dominik Lorenz, Patrick Esser, and Björn
  Ommer.
\newblock High-resolution image synthesis with latent diffusion models, 2021.

\bibitem[Ronneberger et~al.(2015)Ronneberger, Fischer, and
  Brox]{ronneberger2015}
Olaf Ronneberger, Philipp Fischer, and Thomas Brox.
\newblock {U-Net : Convolutional Networks for Biomedical Image Segmentation}.
\newblock \emph{International Conference on Medical Image Computing and
  Computer-Assisted Intervention}, pages 234--241, 2015.

\bibitem[Saharia et~al.(2021)Saharia, Ho, Chan, Salimans, Fleet, and
  Norouzi]{saharia2021image}
Chitwan Saharia, Jonathan Ho, William Chan, Tim Salimans, David~J Fleet, and
  Mohammad Norouzi.
\newblock Image super-resolution via iterative refinement.
\newblock \emph{arXiv:2104.07636}, 2021.

\bibitem[Saharia et~al.(2022{\natexlab{a}})Saharia, Chan, Chang, Lee, Ho,
  Salimans, Fleet, and Norouzi]{saharia2022palette}
Chitwan Saharia, William Chan, Huiwen Chang, Chris Lee, Jonathan Ho, Tim
  Salimans, David Fleet, and Mohammad Norouzi.
\newblock Palette: Image-to-image diffusion models.
\newblock In \emph{ACM SIGGRAPH 2022 Conference Proceedings}, pages 1--10,
  2022{\natexlab{a}}.

\bibitem[Saharia et~al.(2022{\natexlab{b}})Saharia, Chan, Saxena, Li, Whang,
  Denton, Ghasemipour, Ayan, Mahdavi, Lopes, et~al.]{saharia2022photorealistic}
Chitwan Saharia, William Chan, Saurabh Saxena, Lala Li, Jay Whang, Emily
  Denton, Seyed Kamyar~Seyed Ghasemipour, Burcu~Karagol Ayan, S~Sara Mahdavi,
  Rapha~Gontijo Lopes, et~al.
\newblock Photorealistic text-to-image diffusion models with deep language
  understanding.
\newblock \emph{arXiv preprint arXiv:2205.11487}, 2022{\natexlab{b}}.

\bibitem[Sajjadi et~al.(2022)Sajjadi, Meyer, Pot, Bergmann, Greff, Radwan,
  Vora, Lucic, Duckworth, Dosovitskiy, Uszkoreit, Funkhouser, and
  Tagliasacchi]{srt22}
Mehdi S.~M. Sajjadi, Henning Meyer, Etienne Pot, Urs Bergmann, Klaus Greff,
  Noha Radwan, Suhani Vora, Mario Lucic, Daniel Duckworth, Alexey Dosovitskiy,
  Jakob Uszkoreit, Thomas Funkhouser, and Andrea Tagliasacchi.
\newblock {Scene Representation Transformer: Geometry-Free Novel View Synthesis
  Through Set-Latent Scene Representations}.
\newblock \emph{{CVPR}}, 2022.

\bibitem[Salimans and Ho(2022)]{salimans2022progressive}
Tim Salimans and Jonathan Ho.
\newblock Progressive distillation for fast sampling of diffusion models.
\newblock \emph{arXiv preprint arXiv:2202.00512}, 2022.

\bibitem[Schuhmann et~al.(2022)Schuhmann, Beaumont, Vencu, Gordon, Wightman,
  Cherti, Coombes, Katta, Mullis, Wortsman, et~al.]{schuhmann2022laion}
Christoph Schuhmann, Romain Beaumont, Richard Vencu, Cade Gordon, Ross
  Wightman, Mehdi Cherti, Theo Coombes, Aarush Katta, Clayton Mullis, Mitchell
  Wortsman, et~al.
\newblock Laion-5b: An open large-scale dataset for training next generation
  image-text models.
\newblock \emph{arXiv preprint arXiv:2210.08402}, 2022.

\bibitem[Schwarz et~al.(2020)Schwarz, Liao, Niemeyer, and
  Geiger]{schwarz2020graf}
Katja Schwarz, Yiyi Liao, Michael Niemeyer, and Andreas Geiger.
\newblock Graf: Generative radiance fields for 3d-aware image synthesis.
\newblock \emph{Advances in Neural Information Processing Systems},
  33:\penalty0 20154--20166, 2020.

\bibitem[Shen and Zhou(2020)]{shen2020closed}
Yujun Shen and Bolei Zhou.
\newblock Closed-form factorization of latent semantics in gans.
\newblock \emph{arXiv preprint arXiv:2007.06600}, 2020.

\bibitem[Shue et~al.(2022{\natexlab{a}})Shue, Chan, Po, Ankner, Wu, and
  Wetzstein]{shue20223d}
J~Ryan Shue, Eric~Ryan Chan, Ryan Po, Zachary Ankner, Jiajun Wu, and Gordon
  Wetzstein.
\newblock 3d neural field generation using triplane diffusion.
\newblock \emph{arXiv preprint arXiv:2211.16677}, 2022{\natexlab{a}}.

\bibitem[Shue et~al.(2022{\natexlab{b}})Shue, Chan, Po, Ankner, Wu, and
  Wetzstein]{triplanediff}
J.~Ryan Shue, Eric~Ryan Chan, Ryan Po, Zachary Ankner, Jiajun Wu, and Gordon
  Wetzstein.
\newblock 3d neural field generation using triplane diffusion,
  2022{\natexlab{b}}.

\bibitem[Sitzmann et~al.(2019)Sitzmann, Zollh{\"{o}}fer, and
  Wetzstein]{Sitzmann2019}
Vincent Sitzmann, Michael Zollh{\"{o}}fer, and Gordon Wetzstein.
\newblock {Scene Representation Networks: Continuous 3D-Structure-Aware Neural
  Scene Representations}.
\newblock \emph{Advances in Neural Information Processing Systems}, pages
  1119--1130, 2019.

\bibitem[Skorokhodov et~al.(2022)Skorokhodov, Tulyakov, Wang, and
  Wonka]{epigraf}
Ivan Skorokhodov, Sergey Tulyakov, Yiqun Wang, and Peter Wonka.
\newblock Epigraf: Rethinking training of 3d gans.
\newblock \emph{arXiv preprint arXiv:2206.10535}, 2022.

\bibitem[Sohl-Dickstein et~al.(2015)Sohl-Dickstein, Weiss, Maheswaranathan, and
  Ganguli]{sohl2015deep}
Jascha Sohl-Dickstein, Eric Weiss, Niru Maheswaranathan, and Surya Ganguli.
\newblock Deep unsupervised learning using nonequilibrium thermodynamics.
\newblock In \emph{International Conference on Machine Learning}, pages
  2256--2265. PMLR, 2015.

\bibitem[Song et~al.(2020{\natexlab{a}})Song, Meng, and
  Ermon]{song2020denoising}
Jiaming Song, Chenlin Meng, and Stefano Ermon.
\newblock Denoising diffusion implicit models.
\newblock \emph{arXiv preprint arXiv:2010.02502}, 2020{\natexlab{a}}.

\bibitem[Song and Ermon(2019)]{song2019generative}
Yang Song and Stefano Ermon.
\newblock Generative modeling by estimating gradients of the data distribution.
\newblock \emph{Advances in Neural Information Processing Systems}, 32, 2019.

\bibitem[Song et~al.(2020{\natexlab{b}})Song, Sohl-Dickstein, Kingma, Kumar,
  Ermon, and Poole]{song2020score}
Yang Song, Jascha Sohl-Dickstein, Diederik~P Kingma, Abhishek Kumar, Stefano
  Ermon, and Ben Poole.
\newblock Score-based generative modeling through stochastic differential
  equations.
\newblock \emph{arXiv preprint arXiv:2011.13456}, 2020{\natexlab{b}}.

\bibitem[Sun et~al.(2022)Sun, Wang, Shi, Wang, Wang, and Liu]{sun2022ide}
Jingxiang Sun, Xuan Wang, Yichun Shi, Lizhen Wang, Jue Wang, and Yebin Liu.
\newblock Ide-3d: Interactive disentangled editing for high-resolution 3d-aware
  portrait synthesis.
\newblock \emph{arXiv preprint arXiv:2205.15517}, 2022.

\bibitem[Tov et~al.(2021)Tov, Alaluf, Nitzan, Patashnik, and
  Cohen-Or]{tov2021e4e}
Omer Tov, Yuval Alaluf, Yotam Nitzan, Or Patashnik, and Daniel Cohen-Or.
\newblock Designing an encoder for stylegan image manipulation.
\newblock \emph{arXiv preprint arXiv:2102.02766}, 2021.

\bibitem[Wang et~al.(2022{\natexlab{a}})Wang, Du, Li, Yeh, and
  Shakhnarovich]{SJC}
Haochen Wang, Xiaodan Du, Jiahao Li, Raymond~A. Yeh, and Greg Shakhnarovich.
\newblock Score jacobian chaining: Lifting pretrained 2d diffusion models for
  3d generation, 2022{\natexlab{a}}.

\bibitem[Wang et~al.(2022{\natexlab{b}})Wang, Zhang, Zhang, Gu, Bao,
  Baltrusaitis, Shen, Chen, Wen, Chen, and Guo]{rodin}
Tengfei Wang, Bo Zhang, Ting Zhang, Shuyang Gu, Jianmin Bao, Tadas
  Baltrusaitis, Jingjing Shen, Dong Chen, Fang Wen, Qifeng Chen, and Baining
  Guo.
\newblock Rodin: A generative model for sculpting 3d digital avatars using
  diffusion, 2022{\natexlab{b}}.

\bibitem[Wang et~al.(2022{\natexlab{c}})Wang, Zhang, Zhang, Gu, Bao,
  Baltrusaitis, Shen, Chen, Wen, Chen, et~al.]{wang2022rodin}
Tengfei Wang, Bo Zhang, Ting Zhang, Shuyang Gu, Jianmin Bao, Tadas
  Baltrusaitis, Jingjing Shen, Dong Chen, Fang Wen, Qifeng Chen, et~al.
\newblock Rodin: A generative model for sculpting 3d digital avatars using
  diffusion.
\newblock \emph{arXiv preprint arXiv:2212.06135}, 2022{\natexlab{c}}.

\bibitem[Watson et~al.(2022)Watson, Chan, Martin-Brualla, Ho, Tagliasacchi, and
  Norouzi]{watson2022novel}
Daniel Watson, William Chan, Ricardo Martin-Brualla, Jonathan Ho, Andrea
  Tagliasacchi, and Mohammad Norouzi.
\newblock Novel view synthesis with diffusion models.
\newblock \emph{arXiv preprint arXiv:2210.04628}, 2022.

\bibitem[Xu et~al.(2021)Xu, Peng, Yang, Shen, and Zhou]{xu2021volumegan}
Yinghao Xu, Sida Peng, Ceyuan Yang, Yujun Shen, and Bolei Zhou.
\newblock 3d-aware image synthesis via learning structural and textural
  representations.
\newblock 2021.

\bibitem[Yu et~al.(2021{\natexlab{a}})Yu, Ye, Tancik, and Kanazawa]{pixelnerf}
Alex Yu, Vickie Ye, Matthew Tancik, and Angjoo Kanazawa.
\newblock {pixelNeRF: Neural Radiance Fields from One or Few Images}.
\newblock \emph{IEEE Conference on Computer Vision and Pattern Recognition},
  2021{\natexlab{a}}.

\bibitem[Yu et~al.(2021{\natexlab{b}})Yu, Gao, Wang, Yu, Shen, and
  Sang]{yu2021bisenet}
Changqian Yu, Changxin Gao, Jingbo Wang, Gang Yu, Chunhua Shen, and Nong Sang.
\newblock Bisenet v2: Bilateral network with guided aggregation for real-time
  semantic segmentation.
\newblock \emph{International Journal of Computer Vision}, 129:\penalty0
  3051--3068, 2021{\natexlab{b}}.

\bibitem[Zhang and Agrawala(2023)]{zhang2023adding}
Lvmin Zhang and Maneesh Agrawala.
\newblock Adding conditional control to text-to-image diffusion models, 2023.

\bibitem[Zhang et~al.(2018{\natexlab{a}})Zhang, Cui, Neumann, and
  Chen]{zhang2018}
Muhan Zhang, Zhicheng Cui, Marion Neumann, and Yixin Chen.
\newblock {An End-to-End Deep Learning Architecture for Graph Classification}.
\newblock \emph{AAAI}, 2018{\natexlab{a}}.

\bibitem[Zhang et~al.(2018{\natexlab{b}})Zhang, Isola, Efros, Shechtman, and
  Wang]{Zhang2018_lpips}
Richard Zhang, Phillip Isola, Alexei~A. Efros, Eli Shechtman, and Oliver Wang.
\newblock {The Unreasonable Effectiveness of Deep Features as a Perceptual
  Metric}.
\newblock \emph{IEEE Conference on Computer Vision and Pattern Recognition},
  pages 586--595, 2018{\natexlab{b}}.

\bibitem[Zhang et~al.(2018{\natexlab{c}})Zhang, Isola, Efros, Shechtman, and
  Wang]{Zhang2018f}
Richard Zhang, Phillip Isola, Alexei~A Efros, Eli Shechtman, and Oliver Wang.
\newblock The unreasonable effectiveness of deep features as a perceptual
  metric.
\newblock In \emph{CVPR}, 2018{\natexlab{c}}.

\bibitem[Zhou and Tulsiani(2022)]{zhou2022sparsefusion}
Zhizhuo Zhou and Shubham Tulsiani.
\newblock Sparsefusion: Distilling view-conditioned diffusion for 3d
  reconstruction, 2022.

\bibitem[Zhu et~al.(2023)Zhu, Yang, Shen, Shi, Zhao, and Chen]{zhu2023linkgan}
Jiapeng Zhu, Ceyuan Yang, Yujun Shen, Zifan Shi, Deli Zhao, and Qifeng Chen.
\newblock Linkgan: Linking gan latents to pixels for controllable image
  synthesis.
\newblock \emph{arXiv preprint arXiv:2301.04604}, 2023.

\end{thebibliography}
}
\clearpage
\appendix
\section*{\huge Appendix}

\section{Dataset Details}
\lblsec{dataset}
\vspace{-5pt}\paragraph{FFHQ} contains 70k images of real human faces in resolution of $1024^2$. We directly adopted the downsampled, re-aligned version provided by EG3D~\cite{eg3d}, which re-cropped the face and estimate the camera poses.
\vspace{-5pt}\paragraph{AFHQ-cat} contains in total 5K images of cat faces in resolution of $512^2$. The same as FFHQ, we directly download the data with estimated camera poses.
\vspace{-5pt}\paragraph{ShapeNet Cars \& Chairs}
are standard benchmarks for single-image view synthesis~\cite{Sitzmann2019}. We use the data modified by pixelNeRF~\cite{pixelnerf}~\footnote{\url{https://github.com/sxyu/pixel-nerf}}. 
The chairs dataset consists of 6591 scenes, and the cars dataset has 3514 scenes, both with a predefined train/val/test split. Each training scene contains $50$ posed images taken from random points on a sphere. Each testing scene contains $250$ posed images taken on an Archimedean spiral along the sphere. All images are rendered at a resolution of $128^2$. 

\vspace{-5pt}\paragraph{CelebA-HQ Dataset~\cite{Karras2017}} is comprised of 30,000 high-resolution images, each with dimensions of $1024^2$ pixels. For the Seg-to-3D task, we utilize camera poses and labels supplied by Pix2Pix3D~\cite{deng20233d}.

\vspace{-5pt}\paragraph{StyleGAN3-synthetic}. Owing to concerns regarding individual consent, we utilize the StyleGAN3~\cite{karras2021} algorithm to synthesize $165$ images that subsequently facilitate qualitative analysis and video production. This methodology adheres to ethical guidelines while effectively enabling the visualization and evaluation of our findings.  We adhere to the same pre-processing procedure utilized by EG3D~\cite{eg3d} for these synthetic images. This approach involves re-centering the faces and estimating the camera positions, thus ensuring a consistent methodology across datasets.

\section{Implementation Details}
\subsection{3D GAN Settings}
\paragraph{Model}
Our method is largely based on EG3D~\cite{eg3d}~\footnote{\url{https://github.com/NVlabs/eg3d.git}} which adopts tri-plane representations to achieve efficient rendering process. We use the same hyper-parameters as stated in the original paper~\cite{eg3d}, where the triplane dimensions are set $3\times256\times256\times32$ for all datasets.
To stabilize training of diffusion models, we constraints the value of triplanes by bounding its values to $(-1,1)$ with $\tanh(.)$.  We set the neural rendering resolution to be $128\times 128$ for FFHQ and AFHQ-cat following a $\times 8$ 2D-upsampler, while $64\times 64$ for ShapeNet Cars and Chairs following a $\times 2$ 2D-upsampler. 
\vspace{-5pt}\paragraph{Training}
We follow similar recipes~\cite{eg3d} for training EG3D models on four datasets. For FFHQ and ShapeNet, we train EG3D from scratch with $\gamma=1$ and $\gamma=0.3$, respectively. We first train FFHQ model at $64\times 64$ resolution for $25$M images, and another $2.5$M images at $128\times 128$. For ShapeNet, we train both datasets with $10$M images. AFHQ-cat is a much smaller set, so we fine-tune the FFHQ checkpoint directly at $128\times 128$ with $\gamma=5$ and data augmentation~\cite{Karras2020ada} for $4.5$M images. We additionally train an EG3D model on CelebAHQ for comparing on \emph{Seg-to-3D} tasks. For this model, we fine-tune from the pre-trained FFHQ checkpoint with cameras provided by~\cite{deng20233d}. 
Both human and cat face models are trained with ``generator pose conditioning (GPC)''.
Moreover, to encourage a smooth learned tri-plane space, we apply an additional regularization over the L2 norm over the tri-plane with weight $\lambda=1$ for all experiments.
We use a batch size of $32$ on $8$ NIVIDA A100 GPUs, and training approximately takes $3$ days for $25$M images.
\vspace{-5pt}
\paragraph{Inference}
The trained EG3D models are used in both 3D diffusion training \& inference. More precisely, we keep the neural renderer (NeRF MLPs + 2D upsamplers, see \cref{fig:teaser} for illustration) as the final stage of the tri-plane diffusion, which renders the denoised tri-plane into images given the camera input. 
To make sure the rendering solely depending on the tri-plane and viewing directions, we adopt the center-camera for GPC, and input the EMA style vector $\vw_\text{avg}$ as well as constant noise to the upsamplers. We did not notice any quality difference by replacing with the average vectors. 

\subsection{3D Diffusion Settings}
\paragraph{Unconditional Model} We use the improved UNet-based architecture~\cite{ronneberger2015,dhariwal2021diffusion} for all of our main experiments of tri-plane space diffusion. In the exploration stage, we also tried different architectures such as Transformers~\cite{peebles2022scalable}, however, we did not notice significant difference on generation, and keep UNet as the basic backbone. Since the tri-plane size is fixed across various datasets, we apply exactly the same architecture and hyperparameters for all experiments.
Our initial experiments showed that predicting the noise $\veps$ (default setting as suggested by DDPM~\cite{ho2020denoising}) or the velocity $\vv$~\cite{salimans2022progressive} tend to produce noticeable high-frequency artifacts on the generated tri-planes. We suspect it is due to the tri-plane space is naturally noisier than images, and all our models are trained with the signal $\vz_0$ prediction as presented in~\cref{eq:diff_loss} with $\omega_t=\textrm{Sigmoid}(\log(\alpha_t^2/\sigma_t^2))$.
\vspace{-5pt}
\paragraph{Conditional Model}
The main settings of conditional diffusion models are identical to the unconditional models, except for the interaction module between the conditioning input.
For tasks like \emph{3D inversion}, \emph{3D SR}, \emph{Seg-to-3D} and \emph{Edge-to-3D}, we transform the input into RGB images,  resize the spatial resolution into $256\times 256$. Then we jointly train a UNet-based encoder which has the same number of layers and hidden dimensions as the denoiser. Note that, due to the use of self-attention layers~\cite{dhariwal2021diffusion}, the UNet-based encoder is able to globally adjust the features even the input images are not spatially aligned with the canonical tri-plane space. Additionally, similar to \cite{watson2022novel,gu2023nerfdiff}, we include a cross-attention layer between each self-attention outputs of the encoder and denoiser to strengthen the conditional modeling. On the other hand, for both the \emph{Shape-to-3D} and \emph{Text-to-3D} (with CLIP) tasks, we do not train another encoder, but treating the conditioning as vectors which are linearly transformed and combined with the time-embeddings.
\vspace{-5pt}
\paragraph{Training}
We adopt the same training scheme for all our diffusion experiments including unconditional and conditional cases, which uses AdamW~\cite{loshchilov2017decoupled} optimizer with a learning rate of $2e-5$ and a EMA decaying rate of $0.9999$. 
To encourage our high-resolution denosier to learn sufficiently on noisy tri-planes, we adopt a shifted cosine schedule ($256^2\rightarrow 64^2$) inspired by~\cite{hoogeboom2023simple}.
We train all models with a batch size of $32$ for $500$K iterations on $8$ NVIDIA A100 GPUs. 
\vspace{-5pt}
\paragraph{Conditioning camera}
As pointed out in \cref{sec:conditional}, it is critical to train conditional diffusion models with balanced camera poses, wheres the camera viewpoints from natural images (e.g., FFHQ, AFHQ) are typically biased toward the center view. Unlike training 3D GANs where matching the camera distribution is important for learning the 3D space, we found it crucial to have an unbiased input camera distribution when the 3D space is already learned. Otherwise, the performance of conditional generation degenerates heavily when the input image is not center-aligned. Therefore, for human and cat faces, we re-sample the input cameras which looks at the origin and distributes uniformly. To simulate errors in camera prediction, we augment the intrinsic matrix (focal length, $f_x$, $f_y$) with random Gaussian noises.
We do not perform resampling and directly use the training set cameras for ShapeNet as it already covers all viewpoints uniformly.
\vspace{-5pt}
\paragraph{Sampling}
Due to the requirements of proper score function $\ell(.,.)$, we only explored guided diffusion for 3D inversion and supper-resolution, while for the remaining tasks, we use the standard sampling strategy.
No classifier-free guidance~\cite{ho2022classifier} is applied.
By default, the standard ancestral sampling~\cite{ho2020denoising} takes $250$ denoising steps for all of the experiments. 
For 3D inversion, we choose $\ell(.,.)$ to be VGG loss~\cite{zhang2018} with $w_t=7e5\cdot\sigma_t$ in \cref{eq:guide}. We notice that it is essential to use a large decreasing weight to take effective guidance.
For supper-resolution tasks, we use exactly the same objective for guidance, while the loss is computed after down-sampling the rendered image into the input resolution.
For cases using Langevin correction, we additionally apply $10$ correction steps as described in \cref{eq:langevin} where $\delta=0.25$. We only add Langevin steps for the first $50$ denoising steps to save computational cost. The Langevin correction steps are particularly useful for unconditional models.

\subsection{Application Details}
\paragraph{Image-to-3D Inversion} In this task, we independently and randomly select 1,000 images from both the FFHQ and AFHQ datasets, with the results presented in the main paper. To enhance the experimental rigor, we additionally choose 1,000 random images from the test set of CelebA-HQ dataset. We follow the EG3D methodology to re-crop the face and estimate the camera pose for enhanced processing. The results of CelebA-HQ are presented in Table~\ref{table:inversion_celeb}. We select 5 camera poses with yaw angles of -35°, -17°, 0°, 17°, and 35°, and a roll angle of 0° to generate novel view images. The generated images are employed to compute the Fréchet Inception Distance (nvFID)  to the original dataset and the ID metric (nvID) in relation to the input image. 
\vspace{-5pt}\paragraph{Seg-to-3D} Following a recent work (Pix2Pix3D~\cite{deng20233d}), in the Seg2Face process, we randomly select $500$ images from the CelebA-HQ dataset, accompanied by their segmentation maps, and generate $10$ images per input label using different random seeds. Subsequently, we predict the segmentation map for each generated image using a pretrained face-parsing network~\cite{yu2021bisenet}.In the Seg2Cat task, we employ a similar setting. The main distinction lies in the segmentation prediction process. We use the labels from Pix2Pix3D to train the parsing network and subsequently apply it to predict labels from the generated images.
We evaluate the performance by calculating the mean Intersection over Union (MIOU) and average pixel accuracy (MPA) between the input labels and the predicted labels from the generated images. The Fréchet Inception Distance (FID) is computed between the generated images and all images in the CelebAHQ dataset. Single Generation Diversity (SG Diversity) is obtained by measuring the LPIPS metric between each pair of generated images, given a single conditional input.
\vspace{-5pt}\paragraph{Edge-to-3D} We extract the edges for all datasets using informative drawing~\cite{chan2022learning_drawingline}~\footnote{\url{https://github.com/carolineec/informative-drawings.git}}.
\vspace{-5pt}\paragraph{Shape-to-3D} We employ the FLAME template model~\cite{li2017learning_flame} to represent facial shapes and utilize DECA~\cite{feng2021learning_deca} for extracting the corresponding FLAME parameters.
\vspace{-5pt}\paragraph{Text-to-3D} For this task, we utilize CLIP~\cite{radford2021learning_clip} to extract image and text features. During the training phase, we employ the image features, while in the testing phase, we directly use the text features. While it is commonly known that the text and image spaces of CLIP are not fully aligned~\cite{ramesh2022hierarchical}, we find the conditioning is effective as long as both features are normalized before diffusion.
\subsection{Baseline Details}
\vspace{-5pt}\paragraph{GAN Inversion}
Our primary focus is to compare our approach with prevalent 2D GAN inversion methods, such as the direct optimization scheme introduced by \cite{karras2020analyzing_inversion}, which inverts real images into the $\mathcal{W}$ space. Additionally, we examine a related method that extends to the $\mathcal{W+}$ space~\cite{abdal2019image2stylegan_wplus} and directly optimizes the tri-plane, denoted as $\emph{Tri.}$. The implementation is based on EG3D-projector~\footnote{\url{https://github.com/oneThousand1000/EG3D-projector}}.
We initialize all methods with the average $w$ derived from the dataset. For the optimization process, we employ the LPIPS loss~\cite{Zhang2018_lpips} and utilize the Adam optimizer~\cite{kingma2014adam}, conducting 400 optimization steps for each image.
Additionally, we utilize the encoder proposed by \cite{ko20233d_gan_inversion} to directly estimate the $w$ values from images. We employ their pretrained model.
\vspace{-5pt}\paragraph{Pix2Pix3D~\cite{deng20233d}}
We directly utilize the pretrained checkpoints provided by authors~\footnote{\url{https://github.com/dunbar12138/pix2pix3D}}.
\vspace{-5pt}\paragraph{Pix2NeRF~\cite{pix2nerf}} We utilize the values provided by the authors for our analysis. However, due to the absence of released models and quantitative results, our comparison is limited to the ShapeNet chair dataset.
\section{Additional Quantitative Results}
\setlength{\tabcolsep}{2pt}
\begin{table}[tb!]
\begin{center}
\small
\caption{Quantitative comparison on inversion. }
\label{table:inversion_celeb}
\begin{tabular}{llcccccc}
\toprule
&& \multicolumn{6}{c}{CelebA-HQ}\\
\midrule
&&PSNR $\uparrow$ &SSIM $\uparrow$ &LPIPS$\downarrow$  &ID$\uparrow$
&nvFID  $\uparrow$ &nvID$\uparrow$\\
\midrule
{\multirow{3}{*}{Opt.}}
&$\mathcal{W}$&14.98  &0.65 &0.42 &0.54 & 60.67 & 0.50 \\
&$\mathcal{W+}$&16.62 &0.71 &0.34 &0.74 & 51.23 & 0.66  \\
&\emph{Tri.}   &17.52 &0.76 &{\bf 0.12}&{\bf0.92}&185.6&0.50\\
\multirow{1}{*}{Pred.}
&$\mathcal{W}$&14.55 &0.59 &0.54 &0.28& 68.66 &0.26\\
\midrule
 \multicolumn{2}{c}{Ours} &{\bf21.86} & {\bf0.78} & 0.26&0.82& {\bf27.76}& {\bf0.72}
\\
\bottomrule
\vspace{-20pt}
\end{tabular}
\end{center}
\end{table}

\paragraph{Quantitative Results on ShapeNet}
We include additional quantitative results for ShapeNet Cars \& Chairs in~\cref{tbl:shapenet}.
For both cases, we follow the standard evaluation protocol which takes a fixed input view (typically view $64$) as input control, and render from all other cameras. Evaluation is conducted on the test sets. As the results shown in \cref{tbl:shapenet}, while the proposed approach significantly improves over the existing 3D-GAN inversion baselines, and achieves high scores on perceptual scores such as LPIPS and FID, it has a clear gap compared to PixelNeRF in term of PSNR. 
The primary reason for this discrepancy is that PixelNeRF utilizes multi-view supervision during training, whereas our method relies solely on single-view information. Consequently, PixelNeRF can achieve improved performance in certain aspects. In contrast, our GAN-based approach demonstrates both enhanced 3D consistency and sharper outputs, which contribute to the lower FID and LPIPS scores.
\paragraph{Additional Results on CelebA-HQ}
To fully validate the generality of the proposed method, we conduct additional 3D inversion experiments on out-of-distribution (OOD) face data. As shown in \cref{table:inversion_celeb}, we directly apply the model trained from the FFHQ tri-plane space onto CelebA-HQ, and report the single-view inversion performance. Although tested OOD, the proposed {\model} performs stably and achieves larger gains against standard inversion baselines.
 \begin{figure}[t]
    
    \centering
    \includegraphics[width=0.8\linewidth]{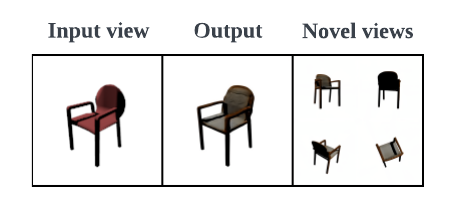}
    \vspace{-10pt}
    \caption{One failure case for conditional generation tasks on ShapeNet Chairs. While {\model} is always able to generate high-fidelity 3D objects, it sometimes fails to recover the texture information from the input view even with guided diffusion.}
    \vspace{-10pt}
    \label{fig:failure_shapenet}
\end{figure}
\section{Additional Qualitative Results}
\label{sec:appendix_results}
\paragraph{3D Inversion \& SR}
We show additional qualitative results of {\model} across datasets for both the 3D inversion (\cref{fig:shapenet_inversion,fig:cat_inversion,fig:face_inversion}) and super-resolution (\cref{fig:superres}) applications.
\begin{figure*}[t]
    \centering
    \includegraphics[width=\linewidth]{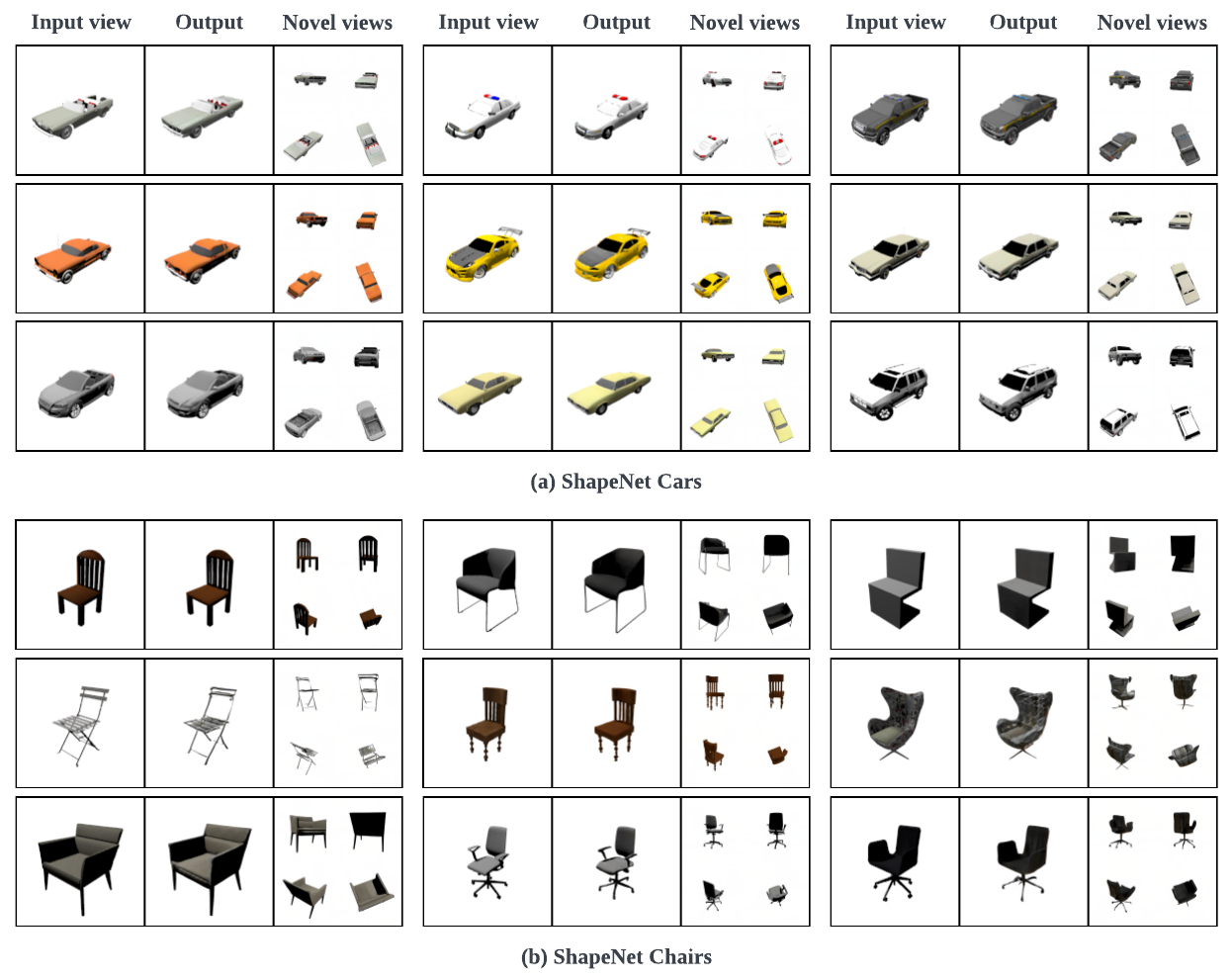}
    \vspace{-20pt}
    \caption{Qualitative results on \textit{3D inversion} for ShapeNet Cars and Chairs. }
    \vspace{-10pt}
    \lblfig{shapenet_inversion}
\end{figure*}
\begin{figure}[t]
    \centering
    \includegraphics[width=0.92\linewidth]{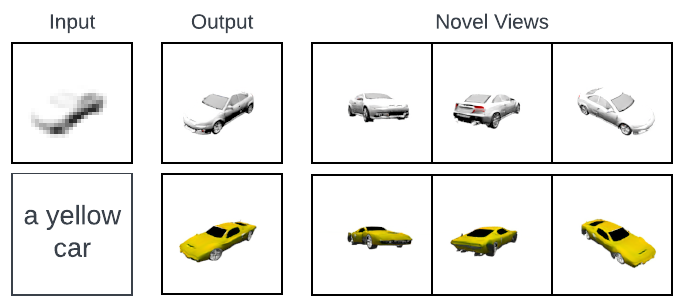}
    \caption{Visual results for super-resolution and text-to-3D tasks on ShapeNet cars.}
    \label{fig:cars}
    \vspace{-15pt}                      
\end{figure}

\captionsetup[subtable]{font=footnotesize}
\begin{table*}[tb!]
 \caption{ \lbltbl{shapenet}
 Quantitative comparison on single-image view synthesis on ShapeNet. *Models have multi-view supervision during training, while our methods including standard optimization-based 3D GAN-inversion baselines are trained with single-view information only.}
\centering
\begin{subtable}{0.4\textwidth}
\centering
\small
\caption{ShapeNet-Cars}
\label{tab:subtable1}
\begin{tabular}{lcccc}
\toprule
& \multicolumn{4}{c}{ShapeNet Cars}\\
\midrule
&PSNR $\uparrow$ &SSIM $\uparrow$ &LPIPS$\downarrow$&FID $\downarrow$\\
\midrule
PixelNeRF~\cite{pixelnerf}$^*$    &{\bf 23.17} & {\bf 0.89} &0.146 & 59.24  \\
3DiM~\cite{watson2022novel}$^*$   &21.01 & 0.57 & –   & 8.99 \\
\midrule
Opt. $\mathcal{W}$ &17.89&0.85&0.124&33.15   \\
Opt. $\mathcal{W+}$&19.23&0.86&0.106&17.95 \\
Opt. \emph{Tri.}  &14.85 &0.63&0.461 &319.8 \\
\midrule
 Ours & 21.13 & {\bf 0.89} & {\bf 0.090} & {\bf 8.86} \\
\bottomrule
\end{tabular}

\end{subtable}
\hspace{1cm}
\begin{subtable}{0.4\textwidth}
\centering
\small
\caption{ShapeNet-Chairs}
\label{tab:subtable2}
\begin{tabular}{lcccc}
\toprule
& \multicolumn{4}{c}{ShapeNet Chair}\\
\midrule
&PSNR $\uparrow$ &SSIM $\uparrow$ &LPIPS$\downarrow$&FID $\downarrow$\\
\midrule
PixelNeRF~\cite{pixelnerf}$^*$    &{\bf 23.72} &{\bf 0.90} &0.128& 38.49  \\
3DiM~\cite{watson2022novel}$^*$    & 17.05& 0.53 &– & {\bf 6.57} \\
\midrule
Pix2NeRF~\cite{pix2nerf}&18.14&0.84&-&14.31\\
Opt. $\mathcal{W}$ &18.28& 0.86&0.110& 10.96   \\
Opt. $\mathcal{W+}$&19.30& 0.87&0.099& 12.70 \\
Opt. \emph{Tri.}  &14.11 & 0.64 & 0.412&237.4 \\
\midrule
 Ours & 20.16 & 0.89 & {\bf 0.090} & 9.76

\\
\bottomrule
\end{tabular}

\end{subtable}
\label{tab:table}
\end{table*}

\begin{figure*}[t]
    \centering
    \includegraphics[width=\linewidth]{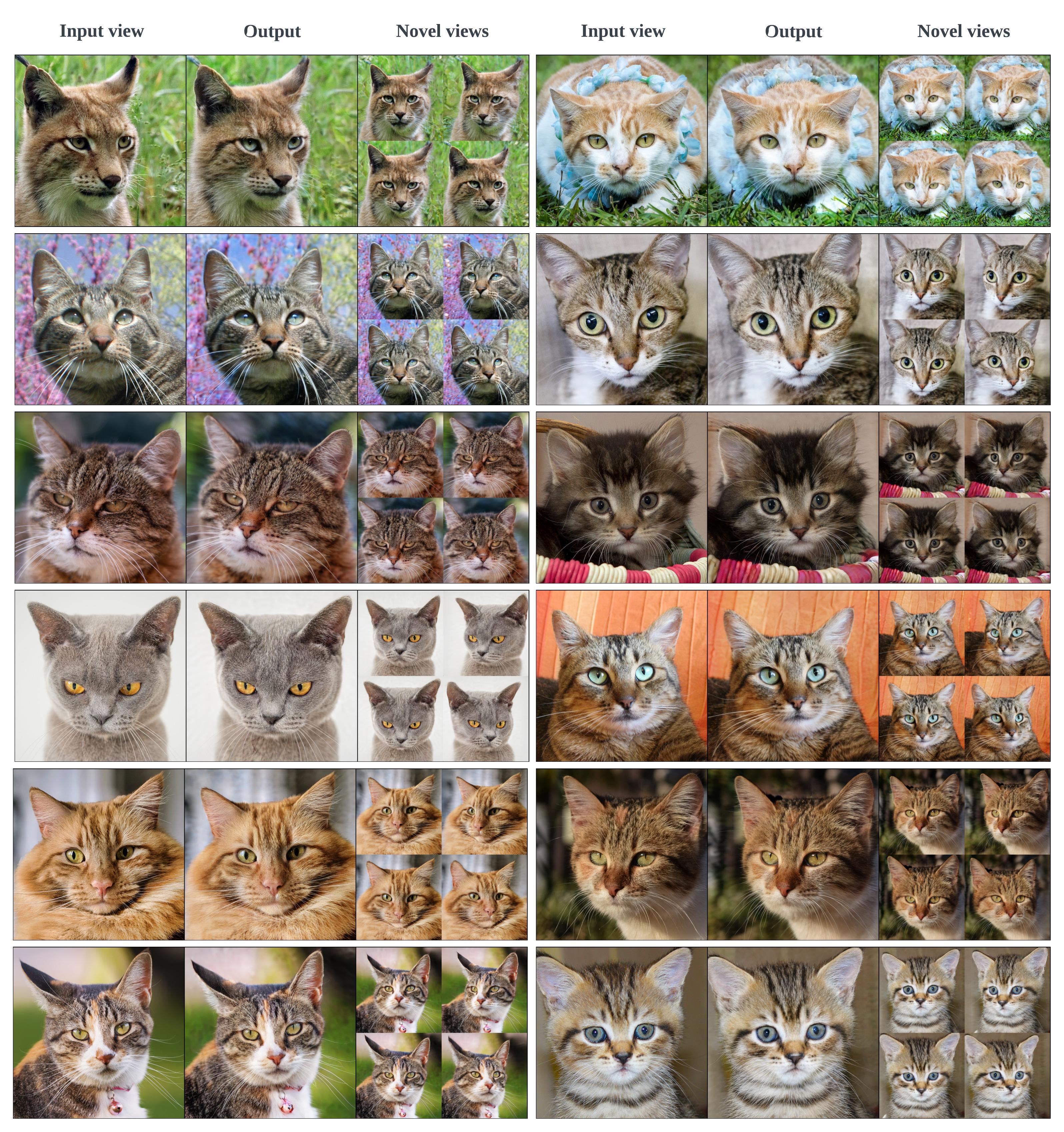}
    \vspace{-20pt}
    \caption{Qualitative results on \textit{3D inversion} for AFHQ-cat. The input images are randomly sampled from the AFHQ training set.}
    \vspace{-10pt}
    \lblfig{cat_inversion}
\end{figure*}

\begin{figure*}[t]
    \centering
    \includegraphics[width=\linewidth]{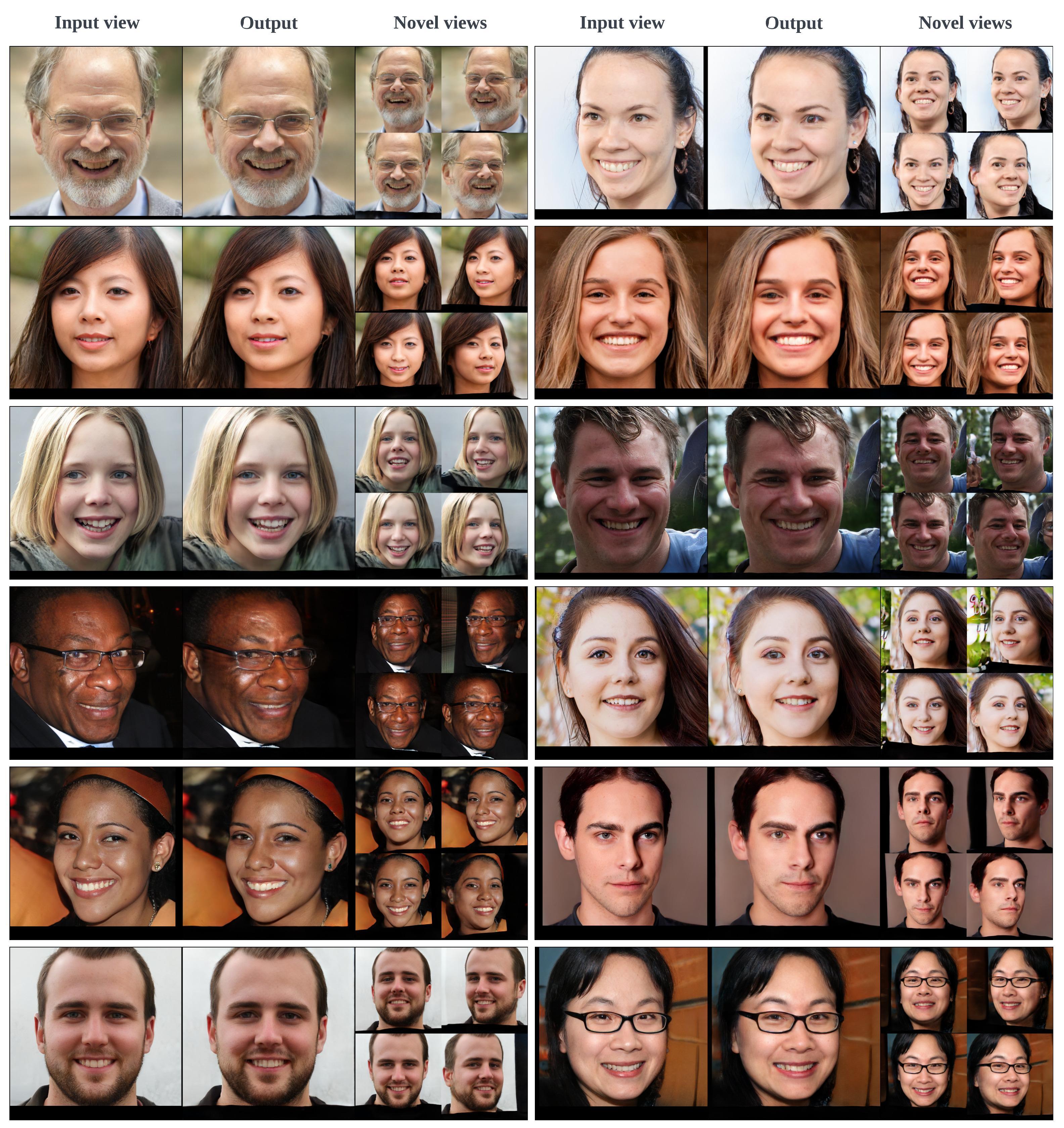}
    \vspace{-20pt}
    \caption{Qualitative results on \textit{3D inversion} for FFHQ. {\bf Due to concerns about individual consent}, all the input faces are synthesized and manually selected from a pre-trained StyleGAN3~\cite{karras2021} checkpoint. We perform exactly the same pre-processing procedure as EG3D~\cite{eg3d} over these synthetic images, which re-centers the faces and estimates the camera positions.}
    \vspace{-10pt}
    \lblfig{face_inversion}
\end{figure*}

\begin{figure*}[t]
    \centering
    \includegraphics[width=\linewidth]{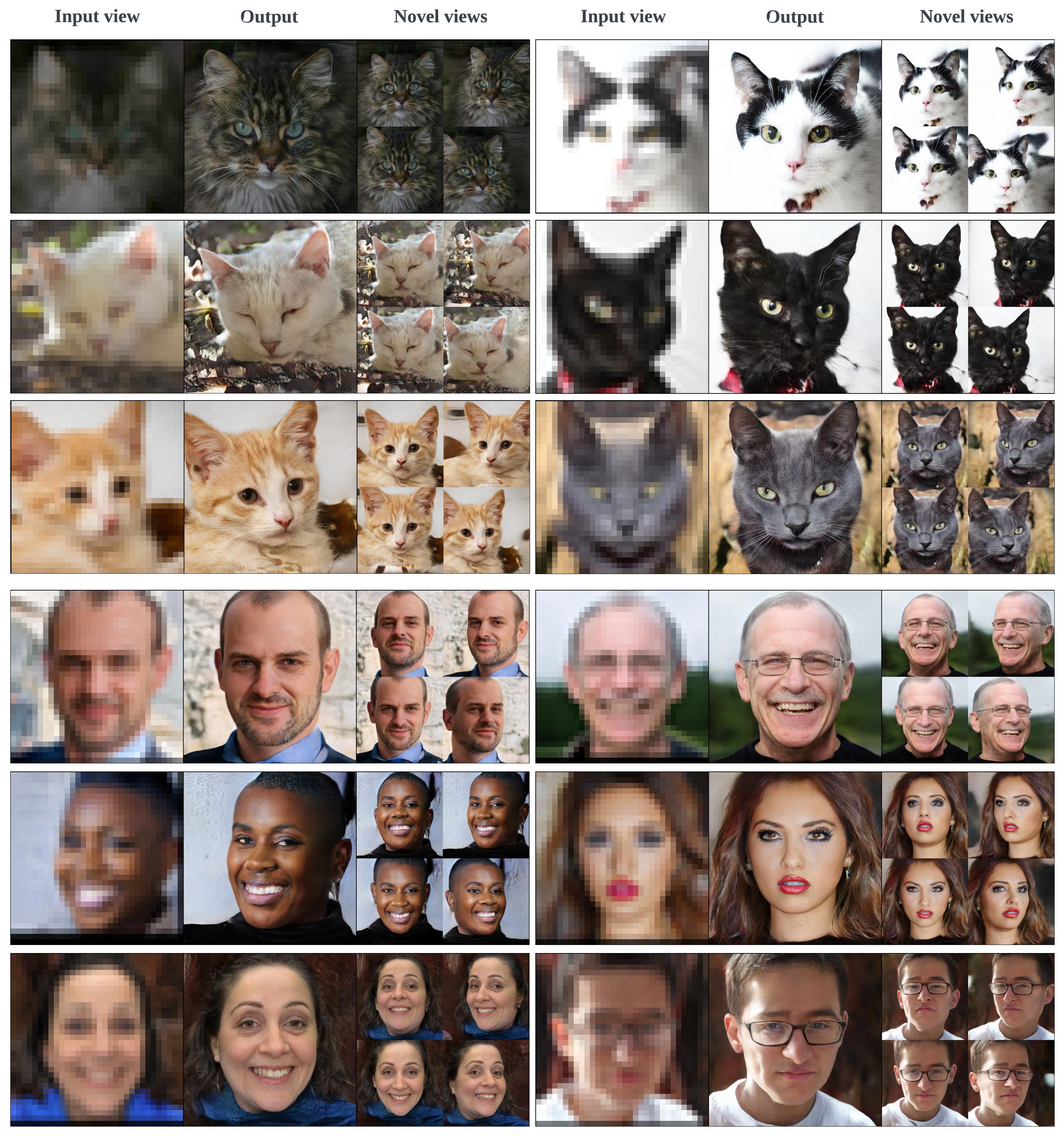}
    \vspace{-20pt}
    \caption{Qualitative results on \textit{3D super-resolution} tasks for AFHQ-cat and FFHQ. }
    \vspace{-10pt}
    \lblfig{superres}
\end{figure*}

\begin{figure*}[t]
    \centering
    \includegraphics[width=\linewidth]{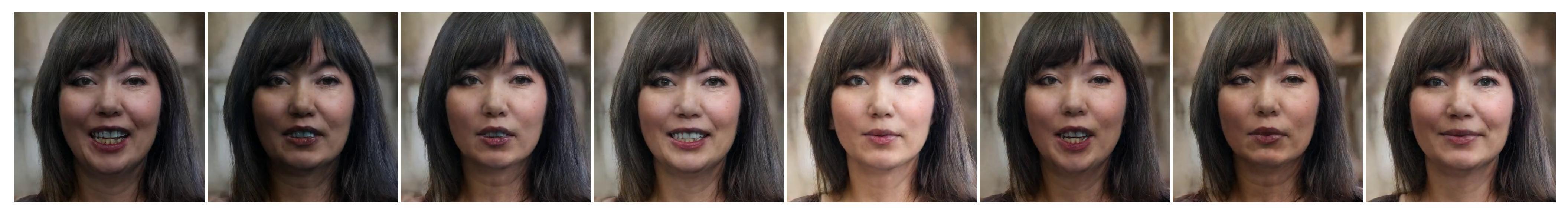}
    \vspace{-20pt}
    \caption{Qualitative results on \textit{Shape-to-3D} for FFHQ. These images can semantically ensure the preservation of identity; however, the color exhibits constant fluctuations. The current control mechanisms are unable to effectively disentangle factors  such as lighting. }
    \vspace{-10pt}
    \lblfig{shape_to_3d}
\end{figure*}

\begin{figure*}[t]
    \centering
    \includegraphics[width=\linewidth]{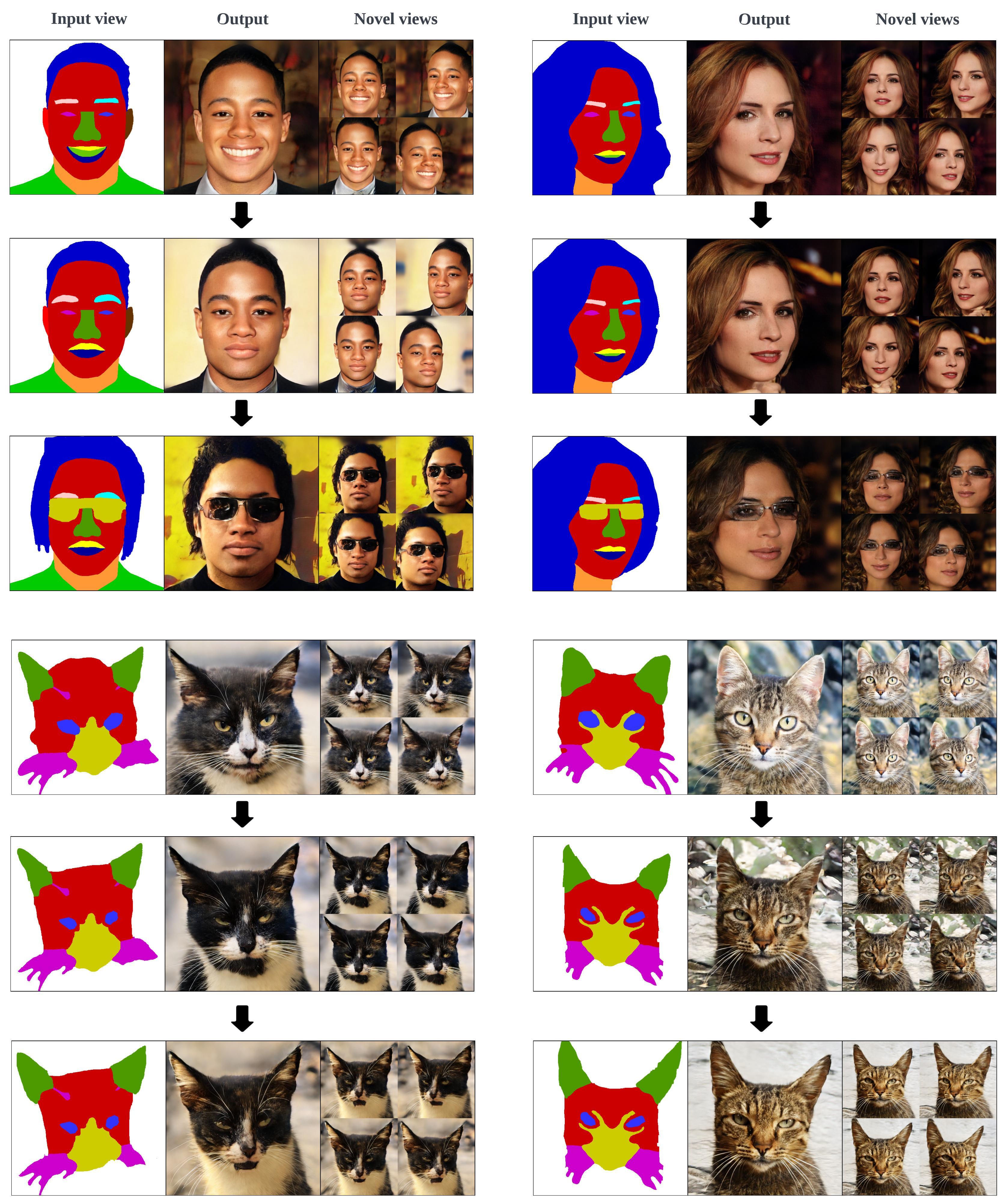}
    \vspace{-20pt}
    \caption{Progressive editing of \emph{Seg-to-3D} synthesis. The input seg-maps are interactively edited. To achieve that, we fix the initial tri-plane noise and use DDIM~\cite{song2020denoising} to obtain diffusion samples.}
    \vspace{-10pt}
    \lblfig{seg_edit}
\end{figure*}

\begin{figure*}[t]
    \centering
    \includegraphics[width=\linewidth]{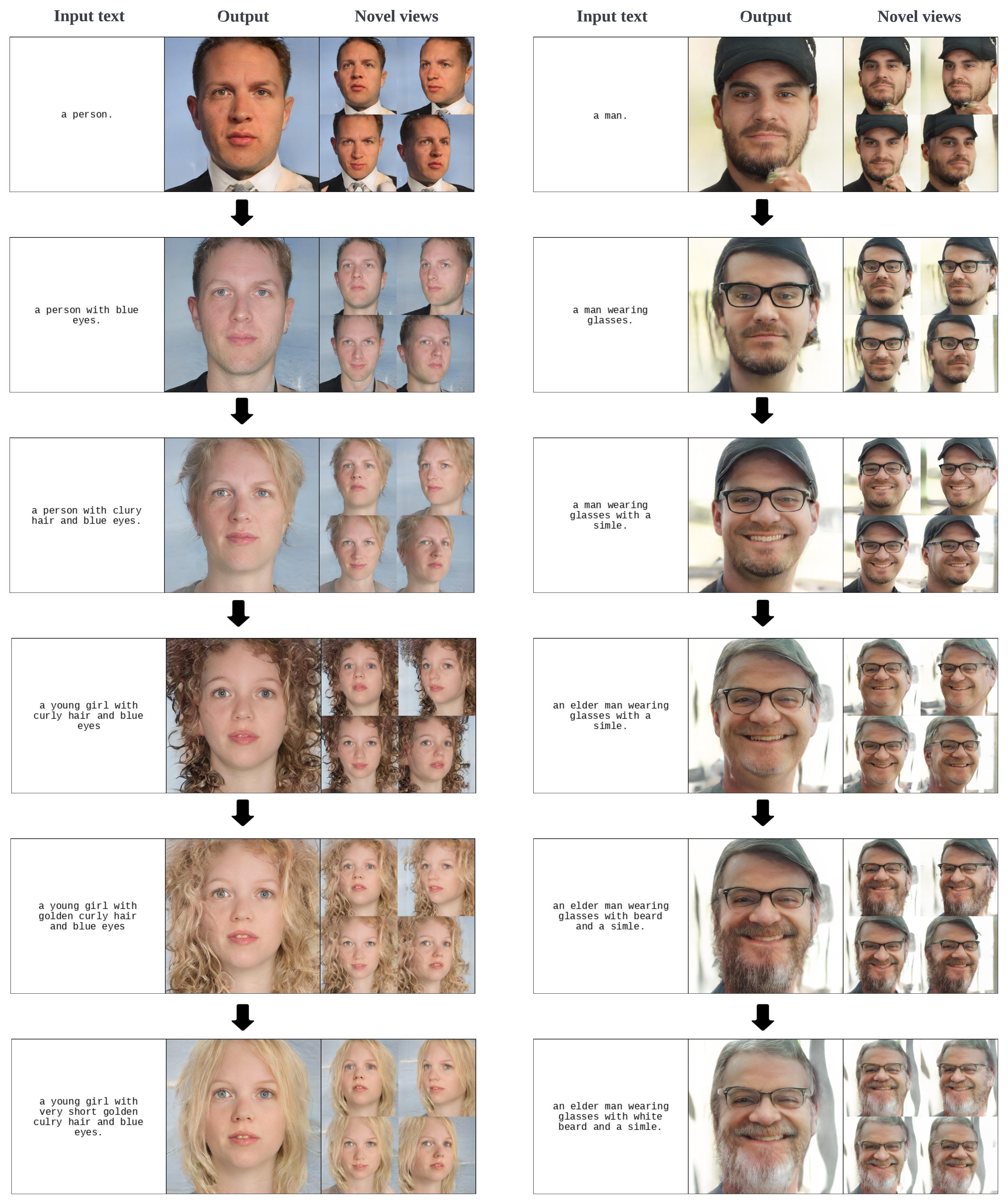}
    \vspace{-20pt}
    \caption{Progressive editing of \emph{Text-to-3D} synthesis. The text prompts will be first transformed to normalized CLIP embeddings, which the diffusion model directly condition on. To achieve that, we fix the initial tri-plane noise and use DDIM~\cite{song2020denoising} to obtain diffusion samples.}
    \vspace{-10pt}
    \lblfig{text_edit}
\end{figure*}

\vspace{-5pt}\paragraph{Seg-to-3D Editing}
\cref{fig:seg_edit} presents an application of our method which supports progressive 3D editing based on 2D segmentation maps.

\vspace{-5pt}\paragraph{Text-to-3D Editing}
The conditional diffusion of {\model} also supports interactive editing given text prompt, as demonstrated in \cref{fig:text_edit}.
\paragraph{Shape-to-3D} \cref{fig:shape_to_3d} presents qualitative results on this task, demonstrating that the generated images can semantically ensure the preservation of identity. However, the color exhibits constant fluctuations. The current control mechanisms are unable to effectively disentangle factors such as lighting. 

\section{Limitations and Future Work}
Our method has two major limitations. 
First, while learning from the latent space of GANs allows us effectively learn controllable diffusion models for 3D, it also brings the drawbacks that GANs commonly have. For instance, a common artifact of the adversarial training is that the learned space typically has mode collapse, which in turn affects the 3D diffusion learning that it may not cover full data space. In our experiments, we particularly noticed this collapsing effect on synthetic datasets with complex geometries such as ShapeNet (see~\cref{fig:failure_shapenet}).
As the future work, this issue can be potentially eased by jointly training the diffusion prior with the 3D-GAN, and including additional image reconstruction loss. 
Moreover, comparing to pure encoder-based approaches~\cite{li20223d}, the iterative nature of the diffusion models generally has a slower generation process.
However, our methods can be easily integrated with existing works for speed-up diffusion models~\cite{salimans2022progressive}. We leave this exploration as future work.


\end{document}